\documentclass[fleqn,usenatbib]{rasti}

\usepackage[utf8]{inputenc} 
\usepackage[T1]{fontenc}    
\usepackage{url}            
\usepackage{booktabs}       
\usepackage{amsfonts}       
\usepackage{nicefrac}       
\usepackage{microtype}      
\usepackage{amsmath}
\usepackage{lipsum}
\usepackage{tabu}
\usepackage{comment}
\usepackage{subfigure}
\usepackage{caption}
\usepackage{makecell}
\usepackage{tabu}
\usepackage{multirow}
\usepackage{graphicx}

\newcommand{\gomez}{G\'{o}mez}

\newcommand{\todo}[1]{\iffalse #1 \fi}

\title[]{\texttt{AnomalyMatch}: Discovering Rare Objects of Interest with Semi-Supervised and Active Learning}

\author[\gomez~et~al.]{
    Pablo \gomez$^{1}$\thanks{E-mail: pablo.gomez@esa.int},
    Laslo E. Ruhberg$^{2}$,
    Maria Teresa Nardone$^{1,3}$,
    David O'Ryan$^{1}$
    \\
    $^{1}$European Space Agency (ESA), European Space Astronomy Centre (ESAC), Camino Bajo del Castillo s/n, 28962, \\ Villanueva de la Cañada, Madrid, Spain\\
    $^{2}$Astronomisches Rechen-Institut, Zentrum f{\"u}r Astronomie der Universit{\"a}t Heidelberg, M{\"o}nchhofstr. 12-14, \\
    69120 Heidelberg, Germany \\
    $^{3}$Kapteyn Astronomical Institute, University of Groningen, Postbus 800, 9700 AV Groningen, The Netherlands
}

\date{Accepted 15/06/2026. Received 06/05/2026; in original form 21/11/2025}

\pubyear{\the\year{}}

\begin{document}
\maketitle
\begin{abstract}
Anomaly detection in large datasets is essential in astronomy and computer vision. However, due to a scarcity of labelled data, it is often infeasible to apply supervised methods to anomaly detection. We present \texttt{AnomalyMatch}, an anomaly detection framework combining the semi-supervised \texttt{FixMatch} algorithm using EfficientNet classifiers with active learning. \texttt{AnomalyMatch} is tailored for large-scale applications and integrated into the ESA Datalabs science platform. In this method, we treat anomaly detection as a binary classification problem and efficiently utilise limited labelled and abundant unlabelled images for training. We enable active learning via a user interface for verification of high-confidence anomalies and correction of false positives. Evaluations on the GalaxyMNIST astronomical dataset and the miniImageNet natural-image benchmark under severe class imbalance display strong performance. Starting from five to ten labelled anomalies, we achieve an average AUROC of $0.96$ (miniImageNet) and $0.89$ (GalaxyMNIST), with respective AUPRC of $0.82$ and $0.77$. After three active learning cycles, anomalies are ranked with 76\% (miniImageNet) to 94\% (GalaxyMNIST) precision in the top 1\% of the highest-ranking images by score. We compare to the established \texttt{Astronomaly} software on selected 'odd' galaxies from the 'Galaxy Zoo - The Galaxy Challenge' dataset, achieving comparable performance with an average AUROC of $0.83$. Our results underscore the exceptional utility and scalability of this approach for anomaly discovery, highlighting the value of specialised approaches for domains characterised by severe label scarcity.
\end{abstract}


\section{Introduction}
Identifying anomalies -- rare and unusual outliers -- in large datasets is a central challenge in machine learning and the applied sciences ~\citep{chandola2009anomaly,Pimentel2014Review}. The term `anomaly' has various definitions dependent upon the field they are being studied in. They can be telemetry outliers, instrumentation effects, objects exhibiting odd morphologies in images or outliers in well-measured distributions. Here we focus on searching for anomalies in image data. Some examples of astrophysical anomalies are gravitational lenses \citep{2022A&A...667A.141G, 2025arXiv250315324E}, galaxy mergers \citep{2022A&A...661A..52P, 2023ApJ...948...40O}, edge-on proto-planetary disks \citep{2024ApJ...967L...3B}, or ring galaxies \citep{2025ApJ...978..137A}. Each of these objects provide unique insights into different aspects of astrophysics.

Anomalies can hold critical relevance as scientifically meaningful discoveries, whether aiding our understanding in galaxy evolution, cosmology or the evolution of objects within the Galaxy. By their nature, anomalies are rare and require large or deep surveys to be detected in large quantities. Next-generation observatories, including \textit{Euclid}~\citep{2011arXiv1110.3193L} and the Vera C. Rubin Observatory~\citep{2019ApJ...873..111I}, are poised to release unprecedented volumes of astronomical data across an unmatched area of the sky. Forecasts for both observatories highlight the scale of the data volumes to be released, with 20 billion galaxies estimated to be observed with the Vera C. Rubin across 18,000 square degrees~\citep{2019ApJ...873..111I} and \textit{Euclid} predicted to observe 1.5 billion galaxies across 14,000 square degrees~\citep[][]{2022A&A...657A..90E, 2025arXiv250315310E} survey area.

Such large surveys provide the only opportunity to study anomalies at a scale where they can be identified to large statistical significance for further study. For example, across the 14,000 square degree footprint of \textit{Euclid}, approximately 170,000 strong gravitational lenses are predicted to be discovered~\citep[roughly 12 per square degree;][]{2025A&A...696A.214P}. Studies across existing archives have also informed us of the incidence rate of such objects. These include \citet[][]{2022A&A...667A.141G}, who identify 252 gravitational lens candidates across the \textit{Hubble} Legacy Archive.

Thus, only surveys of this high depth and over a large area can detect such objects in statistically meaningful numbers. Even rarer are multiply lensed quasars, of which only $\sim$50 were identified across the entire Gaia Data Release 2~\citep{2018A&A...618A..56D}. For more common anomalies, such as galaxy mergers, large surveys offer the opportunity to increase known sample sizes significantly. However, merger catalogues are frequently contaminated by chance-aligned galaxy pairs that are not physically interacting, complicating both detection and classification~\citep{2022A&A...661A..52P, 2023ApJ...948...40O, 2024MNRAS.533.2547F}.

A key challenge in searching anomalies is that, given their rarity (often < 0.1\% of a dataset), there is typically a lack of labelled examples. Existing catalogues may consist of the order of tens of objects, for instance edge on protoplanetary disks \citep{2015ApJ...808L...3A} or quadruply lensed quasars \citep{2021ApJ...921...42S}. In such regimes, even similarity searches struggle to generalise effectively, as the small number of known examples may not capture the full morphological diversity of the class. While similarity searches are powerful complementary tools to anomaly detection methods for expanding known samples \citep[e.g. see][]{2024ApJ...974..172A}, different techniques are required to uncover such objects in the first place. Such techniques must also be efficient enough to identify them in datasets with the scale likely to contain them while remaining robust to the heterogeneity inherent in rare object classes. Due to these needs, machine learning methods are increasingly employed for anomaly detection tasks.

Supervised machine learning methods are most commonly applied to classification tasks in astronomy. Many works have successfully used supervised machine learning to classify astrophysical sources based on their morphology, and expand labelled sets of objects \citep[][]{2023MNRAS.526.4768W}. However, they are often ineffective without large labelled datasets to train them. A promising avenue has been to fine-tune pre-trained neural network models for anomaly detection tasks. Fine-tuning involves taking a model pre-trained on a large dataset and continuing to train it on a new task or different data - e.g. classifying a specific anomaly. This reduces the need for large labelled datasets and training from scratch, but a few hundred examples are often still needed - larger than available catalogues for many types of anomaly - and effectiveness may vary depending on transferability of features and concept drift.

Many works attempt to build the labelled datasets themselves to train supervised methods. Such approaches include tasking citizen scientists \citep[e.g.][]{2013MNRAS.435.2835W, 2a1c3a35730b4eca9c79d201704ebaa5} or have experts visually inspect a dataset to identify candidate anomalies \citep[e.g.][]{2025A&A...697A..14A}. However, it has been found that special care must be taken when instructing individuals to search for different anomalies, or odd objects, as their definition can vary between different citizen scientists and experts \citep{2a1c3a35730b4eca9c79d201704ebaa5}. Another issue is the size of observational datasets to search, as the scale of upcoming surveys makes visual inspection of a large portion of the data impossible.

An approach which can bypass the need for labelled data entirely is that of unsupervised machine learning methods. These methods do not require labelled data at all, and utilise the structure of the underlying data itself in order to identify different classes of objects \citep{scholkopf2001estimating, Zhao2019PyOD}. While effective on lower-dimensional or structured datasets, these methods often struggle with complex high-dimensional data like images, where feature extraction and similarity metrics are more challenging \citep{Goldstein2016Comparative,Zimek2012HighDimensional}.

 However, it is desirable to utilise the known anomalies, even if few, or annotated non-anomalous data when training a model. In many non-astronomical contexts, this has been achieved through semi-supervised learning techniques such as \texttt{MSMatch} \citep{Gomez2021MSMatch}, derived from the well-used \texttt{FixMatch} method \citep{sohn2020fixmatch}. These frameworks combine both supervised and unsupervised methods to learn about the data. \texttt{MSMatch} was able to achieve accuracy competitive with supervised methods on scene classification of satellite imagery with as few as five labelled examples per class while utilising thousands of unlabelled images.

Another approach, combining classical feature space exploration with an element of user-in-the-loop active learning, is that of \texttt{Astronomaly} \citep{Lochner2021Astronomaly}. This algorithm successfully leverages the features of galaxies to break down datasets into different classes, which can then be directed by a user to identify objects of interest. \texttt{Astronomaly} was successfully benchmarked on the Galaxy Zoo: The Galaxy Challenge dataset, a soft label dataset derived from the Galaxy Zoo 2 \citep{2013MNRAS.435.2835W} campaign. They found they were able to recover `Odd' galaxies from a larger dataset with no pre-existing labelled data of the objects.

This approach was further developed in \texttt{Astronomaly: Prot\'eg\'e} \citep[][]{2025AJ....169..121L}, which introduced an additional self-supervised component to the original pipeline. This extension combines feature representations extracted by the self-supervised model with a recommendation system to a user who can then explore the feature space. This combination of a self-supervised approach with active learning facilitates the efficient search for desired anomalies on datasets of up to millions of galaxies.

Examples of algorithms incorporating expert feedback in this iterative way are not limited to imaging data alone, but has also been successfully applied to other astronomical datasets. For instance, the \texttt{coniferest} algorithm \citep[][]{2025A&C....5200960K} was developed by the SNAD team as a dedicated active anomaly detection package operating on tabular feature data. Built upon an IsolationForest, it incorporates active learning components in the form of the Active Anomaly Detection and PineForest algorithms. Applied to successive data releases of the Zwicky Transient Facility, this approach successfully identified a range of astrophysical transients including M-Dwarf flares, supernovae, and RS Canum Venaticorum stars \citep[][]{2021MNRAS.502.5147M, 2023RNAAS...7..155V, 2024MNRAS.533.4309V}.

Such results from the literature show that real-world anomaly detection workflows benefit from directly incorporating human expertise via active learning and user-in-the-loop approaches~\citep{settles2009active}. Such a framework allows iterative model refinement through user interaction, crucially exploiting expert domain knowledge to identify the most valuable anomalies from a scientific perspective. Combining a semi-supervised method - leveraging labelled and unlabelled data - with a user-in-the-loop workflow appears to be key for furthering anomaly detection methods in astronomy.

In this work, we introduce \texttt{AnomalyMatch} - a method that embraces this philosophy. \texttt{AnomalyMatch} is a semi-supervised anomaly detection tool that integrates \texttt{FixMatch} with an EfficientNet-based neural network architecture ~\citep{Tan2019EfficientNet}, active learning, and a custom graphical user interface (GUI). This setup is optimised for anomaly detection as a heavily imbalanced binary classification. Our framework enables domain experts or citizen scientists to iteratively label selected anomalies identified by the model during training, significantly improving detection performance through minimal additional labelling. Originally designed with astronomical datasets in mind, \texttt{AnomalyMatch} is highly generalisable to other domains with similar data characteristics. Moreover, our implementation prioritises scalability: \texttt{AnomalyMatch} can efficiently process and analyse hundreds of millions of images on just one graphics card, as demonstrated in the accompanying article~\citep{HSTPaper}, which makes it particularly suitable for upcoming large-scale astronomical surveys and data repositories. Thus, our method excels in targeted anomaly discovery, empowering users to define and detect anomalies of scientific interest rather than general statistical outliers.

Our primary contributions are:
\begin{itemize}
    \item A novel semi-supervised anomaly detection pipeline (\texttt{AnomalyMatch}) combining \texttt{FixMatch} streamlined for heavy class imbalance, EfficientNet neural networks, and active learning in a heavily optimised framework, effectively leveraging unlabelled data and greatly reducing labelling requirements;
    \item A GUI-based active learning loop enabling experts to iteratively guide anomaly selection and improve model performance during training;
    \item Seamless integration of established astronomy image processing tools, such as different normalisation methods from \texttt{Astropy} \citep{astropy:2013, astropy:2018} and FITS file support.
    \item Comprehensive benchmarking of \texttt{AnomalyMatch} on both daily-life and astronomical image datasets, demonstrating excellent performance in terms of Area Under the Receiver Operating Characteristic (AUROC), Area Under the Precision-Recall Curve (AUPRC), and Anomaly Detection Efficiency with extremely limited labelled samples (five to ten labelled anomalies);
    \item A highly efficient and scalable implementation capable of analysing 100 million images within days ~\citep{HSTPaper}, seamlessly integrated into the European Space Agency's (ESA) Datalabs platform\footnote{\url{https://datalabs.esa.int/}}~\citep{Datalabs};
    \item A detailed discussion on implications for future large-scale anomaly detection applications, including potential expansions to new data modalities, advanced architectures (e.g. vision transformers), and explainable AI.
\end{itemize}

Section~\ref{methods_sect} presents the components of the \texttt{AnomalyMatch} pipeline and provides the reasoning behind its key design choices. Section~\ref{results_sect} describes the extensive testing and benchmarking  of \texttt{AnomalyMatch} on multiple datasets and compares directly to the well used anomaly detection software \texttt{Astronomaly} on a dataset of anomalies from Galaxy Zoo 2. Section~\ref{discussion_sect} discusses the significance of our anomaly detection framework and outlines its assumptions and limitations. Section~\ref{conclusion_sect} includes concluding remarks and discusses potential avenues for future work.

\section{Methods}\label{methods_sect}
\begin{figure*}
    \centering
    \includegraphics[width=0.95\textwidth]{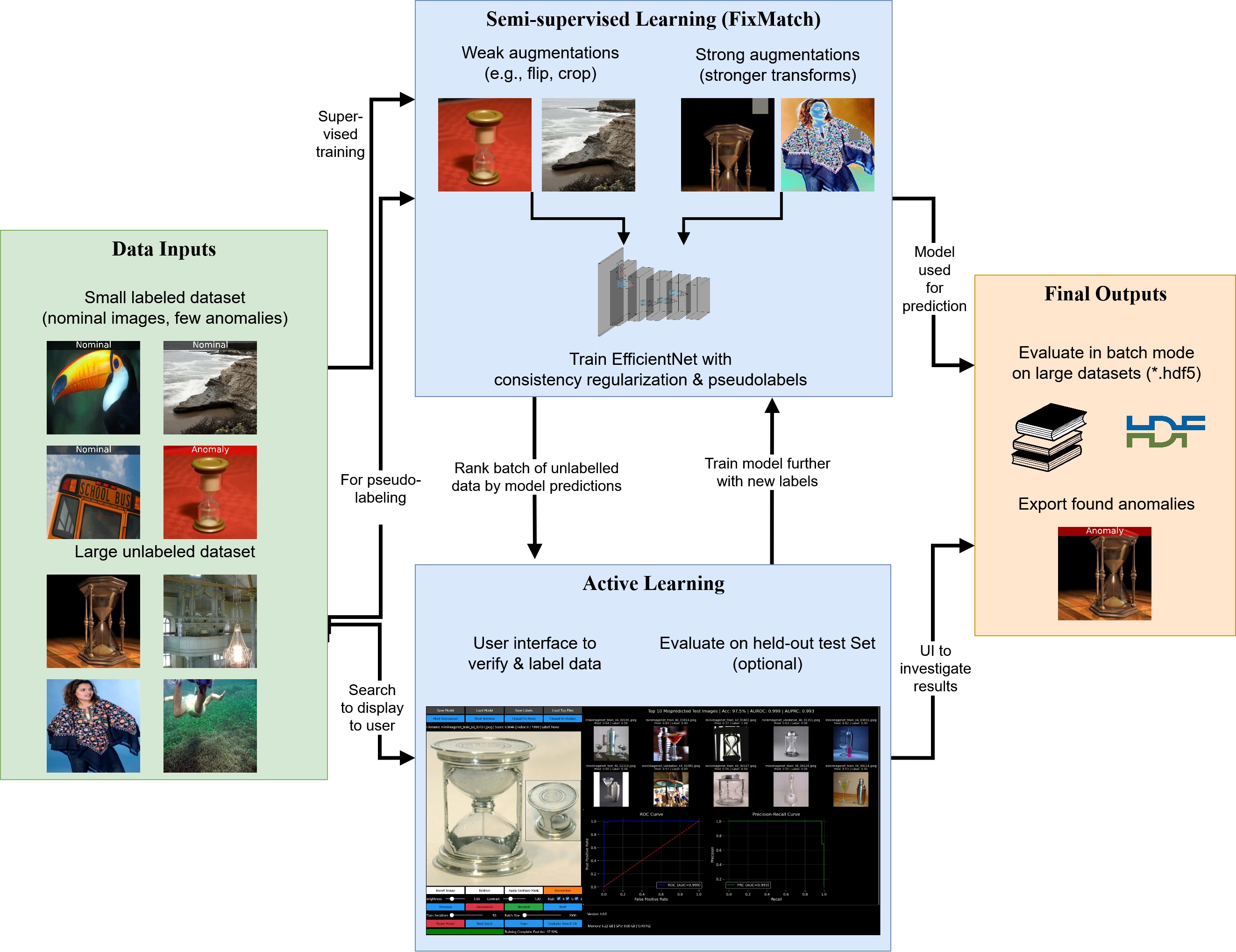}
    \caption{Overview of the \texttt{AnomalyMatch} workflow. A \texttt{FixMatch}-based semi-supervised learning loop trains an EfficientNet backbone using weak and strong augmentations. An active learning interface supports user verification and labelling of additional samples. The final model can be applied in batch mode to large datasets, with detected anomalies exported for further analysis.}
    \label{fig:diagram}
\end{figure*}

Our anomaly detection method, \texttt{AnomalyMatch}, consists of a semi-supervised binary classifier (normal vs. anomaly) built on an EfficientNet backbone fine-tuned for anomaly detection following the \texttt{FixMatch} semi-supervised training strategy. We combine this with an active-learning loop and an intuitive GUI. We detail each component here, highlighting key design decisions. Figure \ref{fig:diagram} shows a summary of the \texttt{AnomalyMatch} methodology which is fully described in this section. It shows the links between the semi-supervised component, active learning component and the expected inputs and outputs of the process.

\subsection{\texttt{FixMatch} for Anomaly Detection}\label{fixmatch_description}
We base our training methodology on \texttt{FixMatch}~\citep{sohn2020fixmatch}, initially proposed for multi-class semi-supervised learning. Underlying the \texttt{FixMatch} strategy is a combined loss function which utilises both labelled and unlabelled data. This loss function is composed of a supervised and unsupervised component. The supervised component is a standard cross-entropy loss, guiding the algorithm using the small labelled dataset that is provided to it. The unsupervised component leverages the structure of the unlabelled data through consistency regularisation.  A weak augmentation is applied to each unlabelled image, and if the model assigns a high-confidence prediction to this weakly augmented version, it is assigned a pseudo-label. A strongly augmented version of the same image is then passed through the model, and the model is penalised if its prediction does not match the pseudo-label. The \texttt{FixMatch} strategy therefore encourages the model to make consistent predictions under different augmentations, effectively learning from the underlying structure of the unlabelled data.

Variants of \texttt{FixMatch} have been successfully applied to specific domains, such as remote sensing imagery~\citep{Gomez2021MSMatch}, semantic segmentation~\citep{marti2022dense, yang2022unimatch}, and graph classification~\citep{GraphixMatch2024}. These adaptations highlight \texttt{FixMatch}’s flexibility in handling various data types and prediction tasks. The success of \texttt{FixMatch} and related approaches across diverse settings motivates their adaptation to anomaly detection tasks characterised by severe label scarcity and class imbalance.

We adapt \texttt{FixMatch} to treat the problem of anomaly detection as a binary classification: "normal" (regular data) vs. "anomaly" (rare, interesting data) with heavy class imbalance. We use a labelled set $\mathcal{D}_L = \{(x_i, y_i)\}$ and an unlabelled set $\mathcal{D}_U = \{u_j\}$, typically with $|\mathcal{D}_U| \gg |\mathcal{D}_L|$ and severe imbalance in favour of normal examples. A practical difference to applying \texttt{FixMatch} in our scenario is that additional labels for normal data are easier to obtain, whereas the anomaly labelled dataset is small and difficult to expand due to their rarity. To handle this severe imbalance, we include oversampling using a weighted random sampler with occurrence-based sampling probabilities of anomalies / normal samples during training.

To train in an unsupervised manner, consistency regularisation is used. Let $f_\theta(x)$ denote the neural network model with parameters $\theta$, outputting the score $p(y=\text{anomaly}|x)$ for input $x$. In each training step, unlabelled images from $\mathcal{D}_U$ undergo weak augmentation (simple flips, crops) to generate pseudo-labels if the model’s predicted probabilities exceed a high-confidence threshold $\tau$ (e.g., 0.95). The same unlabelled images are strongly augmented (applying substantial random transformations) and then fed into the network, enforcing consistency by training the model to match its predictions to the pseudo-labels.

This only comprises half the loss function when training the model. The overall loss combines the unsupervised consistency regularisation with a supervised binary cross-entropy loss calculated using labelled data as
\begin{equation}
L = L_{\text{sup}} + \lambda \cdot L_{\text{unsup}} .
\label{eqn:loss_func}
\end{equation}

\noindent We set $\lambda = 1$ in Equation \ref{eqn:loss_func} to emphasize unlabelled data contributions. Importantly, we find that training for fewer iterations (on the order of hundreds of iterations rather than tens of epochs) is beneficial in the human-in-the-loop setup, as prolonged training sessions risk overfitting to the initial labels. This is explored in more detail in the Section \ref{results_sect}. The \texttt{FixMatch} training strategy is applied to a pre-trained EfficientNet backbone, described in the following subsection, which allows the model to reach high classification accuracy with far fewer training iterations and labelled examples than training from scratch.

\subsection{EfficientNet Backbone}
For computational efficiency and high accuracy, we chose EfficientNetLite0~\citep{Tan2019EfficientNet}. This architecture achieves improved efficiency by systematically balancing depth, width, and resolution of the network through compound scaling as our backbone convolutional neural network uses input images resized to $224\times224$. Smaller input sizes, such as $150\times150$, were also successfully employed in our companion paper searching \textit{Hubble Space Telescope} images ~\citep{HSTPaper}. The network is initialised using ImageNet pre-trained weights, fine-tuning all layers during \texttt{FixMatch} semi-supervised training. Weight decay ($\ell_2$ regularisation) is included to prevent overfitting. The choice of the input image size and normalisation is closely linked to the data augmentation and processing strategy described below.

\subsection{Data Processing and Augmentations}
Robust data augmentation is essential to effective semi-supervised learning. We start from augmentation strategies validated by \texttt{MSMatch}~\citep{Gomez2021MSMatch} and \texttt{DistMSMatch}~\citep{ostman2023decentralised} for the two-tiered augmentation approach:

\begin{itemize}
    \item \textbf{Weak Augmentations.} Minor transformations (horizontal flips, slight random crops) that preserve image content integrity, essential for accurate pseudo-label generation;
    \item \textbf{Strong Augmentations.} Applying RandAugment~\citep{Cubuk2020RandAugment}, we perform aggressive transformations including rotations, colour distortions (brightness, contrast, solarisation), sharpness variations, shearing, translations, and posterisation. These diverse distortions encourage model robustness and invariance to image variability.
\end{itemize}

Each of these augmentations is conducted within \texttt{AnomalyMatch} without input from the user. However, incorporating user feedback is imperative to our approach.

\subsection{Active Learning Loop \& User Interface}
\citet{Lochner2021Astronomaly} demonstrated that incorporating a human in the loop improves the precision of anomaly recovery. We therefore introduce an active learning loop through which the user can iteratively guide and refine the trained model, ensuring that the search remains focused on scientifically meaningful anomalies.

\begin{figure*}
    \centering
    \includegraphics[width=0.99\textwidth]{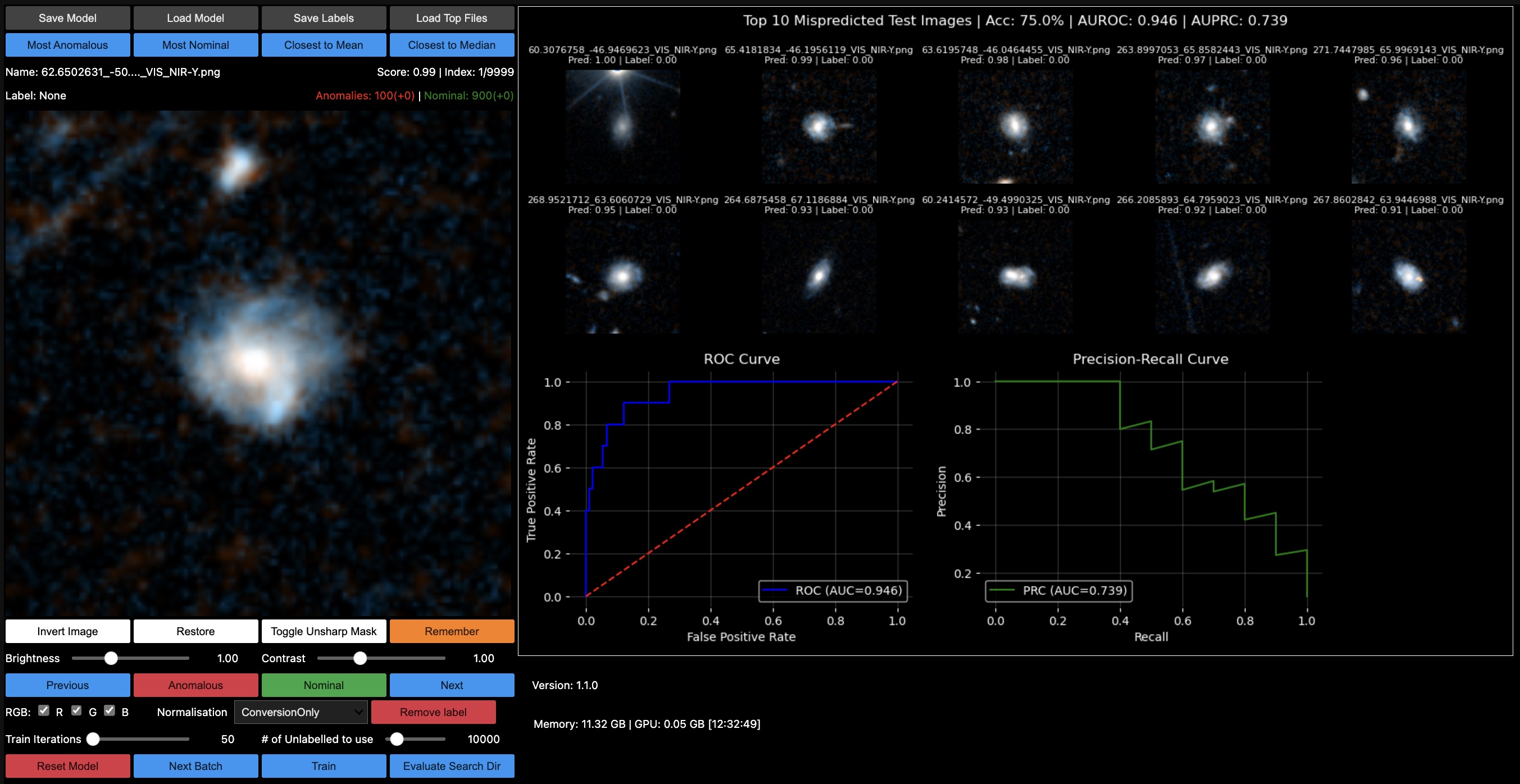}
    \caption{\texttt{AnomalyMatch} active learning interface built with \textit{ipywidgets}. Users can inspect scored unlabelled images and interact with the training set and model through a set of dedicated buttons and controls. Labelling is performed via the `Anomalous' and `Nominal' buttons, which assign a label to the currently displayed image and update the training set accordingly. Visual inspection can be aided by adjusting the `Brightness' and `Contrast' of the image using the sliders and by selecting an image stretch from the `Normalisation' dropdown menu. Training is initiated via the `Train' button, with the number of training iterations and the size of the unlabelled image pool controlled by the `Train Iterations' and `\# Unlabelled to Use' sliders, respectively. Model performance is monitored through displayed AUROC and AUPRC curves, updated after each training cycle. The top mispredicted images from the labelled training set are also displayed to guide further labelling decisions, with the fraction of the training set used for this evaluation set by the user. Finally, the `Evaluate Search Dir' button applies the trained \texttt{AnomalyMatch} model to all images in a specified directory. All model parameters and labels are automatically saved between sessions}
    \label{fig:ui}
\end{figure*}

Leveraging human expertise is central to our pipeline. After initial training, the unlabelled data used is scored and ranked from most to least anomalous. A user is able to then confirm or correct the algorithm, both expanding the training set and refining the model. This also gives initial feedback to the user on the models performance via different metrics (see Appendix \ref{mathematical_definitions} for definitions of all metrics used in this work). Figure~\ref{fig:ui} shows the dedicated interface through which the user conducts the active learning loop.

The GUI provides various practical features:

\begin{itemize}
    \item \textbf{Image Display Tools.} Adjustments for brightness, contrast, unsharp masking, and channel-specific filters (RGB) to support precise labelling;
    \item \textbf{Image Processing Tools.} Adjustments for the normalisation and stretch applied when loading, training and evaluating images;
    \item \textbf{Performance Feedback.} Immediate feedback through AUROC and AUPRC metrics if a test set was defined initially;
    \item \textbf{Interactive Data Management.} Users can load batches of unlabelled data, save session labels and models, and iteratively continue cycles of labelling and retraining.
\end{itemize}

Each of these processes can be performed through the \texttt{AnomalyMatch} GUI without directly interacting with the backend (although this remains possible for advanced users).  The typical workflow, summarised in Figure~\ref{fig:diagram}, would proceed as follows using the GUI shown in Figure \ref{fig:ui}.

The user begins by loading a small labelled dataset which contains both the nominal images and a few anomaly examples and a larger pool of unlabelled images. This forms the two data inputs illustrated in Figure~\ref{fig:diagram}. Training is then initiated via the `Train' button, with the number of training iterations and the size of the unlabelled image pool controlled by the associated sliders. A progress bar appears to show how many training epochs have been completed.

Once training is complete, \texttt{AnomalyMatch} scores the unlabelled pool and displays the highest-scoring image to the user. Alternatively, images can be ordered by their proximity to the median, mean, or lowest score. Active learning then begins and the user visually inspects as many images as they desire. If an anomaly is identified for later inspection using the `Remember' button, the object ID and filename are saved to a separate CSV file. To add the image directly to the labelled training set, the user marks it as anomalous. High-scoring nominal images can similarly be added. In this way, the labelled dataset is iteratively expanded through expert interaction. Once the user has visually inspected as many images as desired, they can train the model again on the expanded training set. This cycle is then repeated as many times as desired.

Model performance is tracked via AUROC and AUPRC curves displayed on the right of the interface, updated after each training cycle provided the user has designated a fraction of the labelled data as a held-out test set. The top mispredicted objects from this test set are also displayed to guide further labelling decisions. Finally, once satisfied with model performance, the user presses `Evaluate Search Dir', providing a path to the full image dataset they wish to classify. All images in this directory are then scored by \texttt{AnomalyMatch}, with results saved as both a CSV and a \texttt{numpy} array (.npy file) containing anomaly scores and associated filenames or source IDs, forming the final outputs shown in Figure~\ref{fig:diagram}. This full process is described in the README of the \texttt{AnomalyMatch} software\footnote{Link to README: \url{https://github.com/esa/AnomalyMatch/blob/main/README.md}.}.

This active learning strategy distinguishes \texttt{AnomalyMatch} from many other anomaly detection approaches in the literature. By keeping a human in the loop and allowing the user to directly guide the model, \texttt{AnomalyMatch} is robust against surfacing uninteresting outliers. For instance, if the user begins to notice that imaging artifacts - such as bad CCDs, defocused regions, or spurious detections~\citep[for examples, see][]{2026NewA..12202466S} - are being scored highly, they can easily add these examples to the labelled dataset as nominal objects. Due to the semi-supervised nature of \texttt{AnomalyMatch}, such artifacts are rapidly assigned low anomaly scores after only one or two labelled examples are provided.

While the intuitive GUI allows for direct interaction with the data, and visual inspection of the model outputs, they only provide a guide and do not require all images to be visually inspected. Instead, the \texttt{AnomalyMatch} scores provide a ranking that can be used for further discovery and provide interesting subsets of the data for further identification. While this could raise concerns about scalability, the semi-supervised nature of \texttt{AnomalyMatch} ensures that even with limited guidance from the training set, the model quickly learns to suppress the score of unwanted nominal objects and converges on the desired anomaly class.

The user-friendly interface shown in Figure \ref{fig:ui} leverages \textit{ipywidgets}, facilitating interactive exploration within Jupyter notebooks. A prebuilt setup integrated within ESA Datalabs further provides seamless access to large-scale astronomical datasets, such as the \textit{Euclid} Quick Data Release~\citep[Q1;][]{aussel2025euclid}. This enables immediate practical use in scientific analyses.

\subsection{Implementation and Scalability}
Our pipeline is built to ensure scalability, robustness, and reproducibility. Initially derived and extensively refactored from an existing PyTorch implementation of \texttt{FixMatch}\footnote{\url{https://github.com/LeeDoYup/FixMatch-pytorch} -- PyTorch code for \texttt{FixMatch} (Accessed 2025-03-28)}, our architecture has been optimised for large-scale datasets, supporting efficient handling of datasets on terabyte scale~\citep[e.g. 100 million \textit{Hubble} images][]{HSTPaper}.

To maintain efficiency over small and large datasets, \texttt{AnomalyMatch} is able to process the following image formats: 

\begin{itemize}
    \item Single-channel or RGB images in PNG or JPEG format;
    \item Arbitrary-channel images stored as TIFF files;
    \item Astronomical images in FITS format.
\end{itemize}

For large scale applications involving thousands to millions of images, we recommend storing these files hierarchically in HDF5 or Zarr formats~\citep{folk2011overview}. These are not image formats themselves, but are storage containers optimised for efficient organisation and fast read access at scale. \texttt{AnomalyMatch} can open these directly, and read the images for classification efficiently. Additionally, the latest \texttt{AnomalyMatch} versions can be run directly over source catalogues from \textit{Euclid} rather than images. This is achieved by integrating the new ESA cutout generator \texttt{Cutana} \citep[][]{2025arXiv251104429G} into \texttt{AnomalyMatch}. In the case of FITS files \texttt{AnomalyMatch} can specify one or 3 extensions to use. Furthermore, the user can select between different normalisation options such as the simple conversion, logarithmic or Asinh stretching or a Z-Scale based linear stretch.

Important hyperparameters of \texttt{AnomalyMatch} include the number of unlabelled images per training cycle and the input image resolution, both of which directly influence training efficiency and detection ability. The number of unlabelled images per training iteration is constrained by available GPU memory. Similarly, the total number of unlabelled images plays an important role. Using too many unlabelled images will result in a large fraction of the unlabelled dataset being rarely seen by the model, eliminating their contribution to training. The input image resolution is equally important, as images are downsampled to a fixed resolution for the EfficientNet backbone. If the resolution is set too low, morphological features critical for anomaly classification may be lost, potentially degrading detection performance.

Continuous integration and deployment workflows incorporating over 70 unit tests ensure code robustness, reliability, and maintainability. This is essential given the complex interaction of semi-supervised and active learning components. The complete source code of \texttt{AnomalyMatch} is publicly available on GitHub\footnote{\url{https://github.com/esa/AnomalyMatch}}, encouraging community collaboration and ongoing development.

\section{Data \& Experimental Setup}
\begin{figure*}
    \centering
    \includegraphics[width=0.99\textwidth]{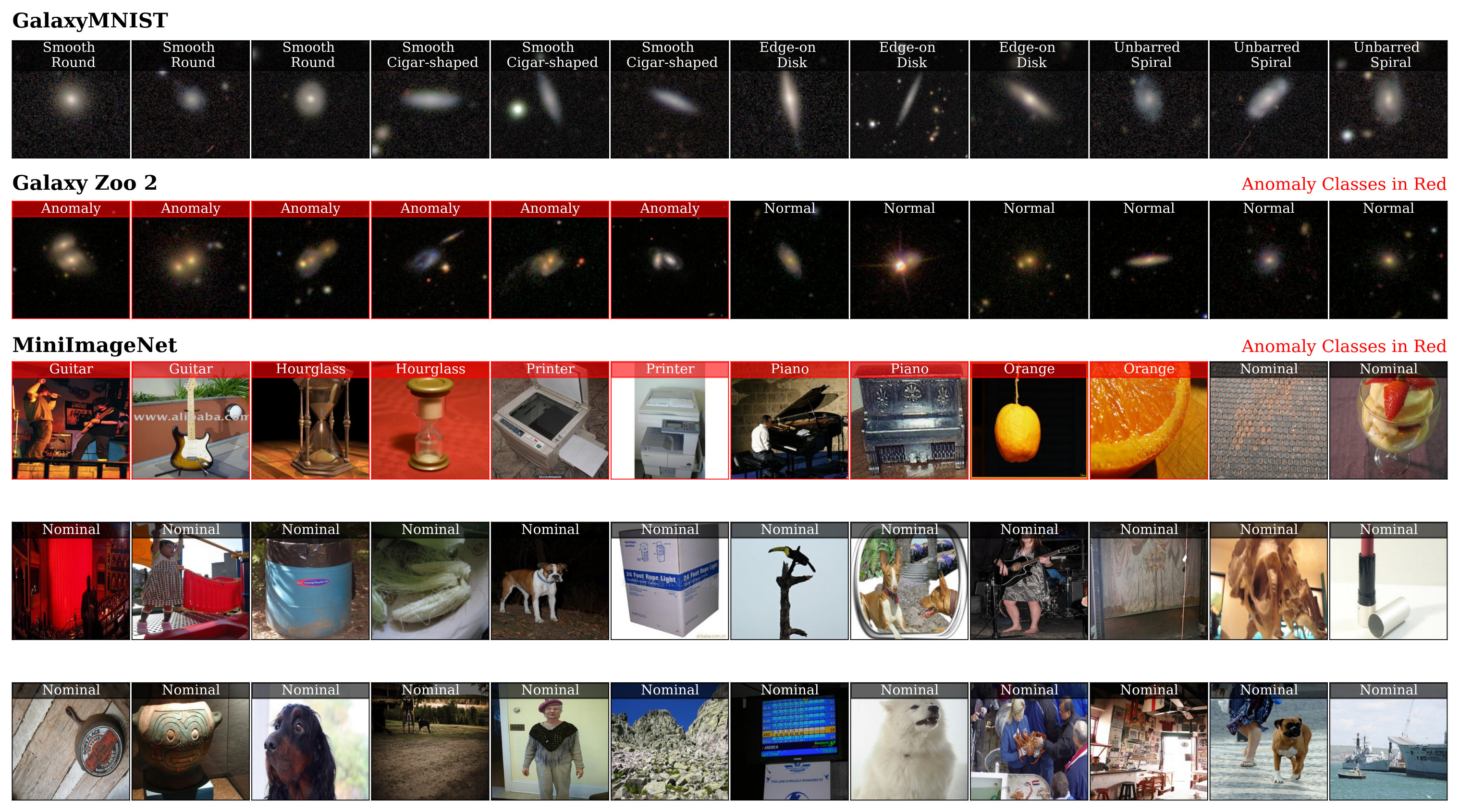}
    \caption{Sample images from GalaxyMNIST (top), Galaxy Zoo 2 (centre), and MiniImageNet (bottom). Red borders indicate anomaly classes used during evaluation, while normal samples are shown with white borders. For GalaxyMNIST all classes were tested as anomaly class in a dedicated run.}
    \label{fig:dataset_samples}
\end{figure*}

We evaluate \texttt{AnomalyMatch} on three datasets: miniImageNet~\citep{Vinyals2016MatchingNetworks}, a common benchmark used in machine learning research; GalaxyMNIST~\citep{walmsley2022galaxy}, a dataset pertinent to our primary application area, astronomy; and a subset of $\approx 65,000$ galaxy images extracted from Galaxy Zoo 2~\citep{2013MNRAS.435.2835W} which were used for the 'Galaxy Challenge' on Kaggle\footnote{\url{https://www.kaggle.com/c/galaxy-zoo-the-galaxy-challenge}}. The initial configurations (label counts and anomaly ratios) varied by dataset. Detailed descriptions of each dataset, along with their respective initial configurations, are provided in the following Sections. Figure~\ref{fig:dataset_samples} shows representative examples from both datasets.

The \texttt{FixMatch} algorithm~\citep{sohn2020fixmatch} formed the basis of our semi-supervised training procedure. Key hyperparameters included an exponential moving average (EMA) momentum of 0.99, confidence threshold ($\tau$) of 0.95, temperature of 0.5, unsupervised-to-supervised loss ratio ($\lambda$ in Equation \ref{eqn:loss_func}) of 1.0, a batch size of 16, learning rate of 0.0075 (with a stochastic gradient descent (SGD) optimiser, momentum 0.9), and weight decay of $7.5\times10^{-4}$. We followed empirical results from \citet{Gomez2021MSMatch,ostman2023decentralised} in the selection of parameters.

To simulate active learning for all three datasets we performed three consecutive training cycles of 100 iterations each. At the end of each cycle, the 10 highest-scoring anomalies and 10 highest-scoring false positives (normal images with high anomaly classification scores by the model) were labelled and incorporated into the training set for the next cycle. The models were finally evaluated on all unlabelled images. 

Throughout the experiments, we employed three key performance metrics: AUROC, AUPRC, and Anomaly Detection Efficiency. AUROC and AUPRC provide complementary views of classification performance, especially important under severe class imbalance. The Anomaly Detection Efficiency quantifies the fraction of anomalies correctly identified within a specified percentage of top-scoring samples, directly reflecting practical usefulness in scenarios involving human-in-the-loop validation and limited labelling resources. The mathematical definitions of all three metrics are given in Appendix~\ref{mathematical_definitions}.

\subsection{miniImageNet}
miniImageNet~\citep{Vinyals2016MatchingNetworks} is a widely used benchmark in few-shot learning. It consists of colour images grouped into 100 classes and is taken from the larger ImageNet dataset. Each class contains 650 images, for a total of 65,000 images in the dataset. miniImageNet was specifically chosen as it is computationally more efficient than the full ImageNet dataset, making it well-suited for our experimentation. Each image was resized from its original resolution to $224\times224$ pixels.

We deliberately selected five distinct object classes as anomalies to evaluate our model's performance across diverse visual features: \textit{Guitar}, \textit{Hourglass}, \textit{Printer}, \textit{Piano}, and \textit{Orange}. As only one class was the anomaly, this represented searching for an object which represented 1\%. Initial labelled data included 500 images (5 anomalies, 495 normal). Each training step utilised an additional randomly sampled pool of 10,000 unlabelled images. These unlabelled images are used in the training of the \texttt{AnomalyMatch} model in the semi-supervised method as described in Section \ref{fixmatch_description}. They are then ordered by score, and the top-scoring objects can be added to the labelled dataset if the user wishes.

\subsection{GalaxyMNIST}
GalaxyMNIST~\citep{walmsley2022galaxy} comprises annotated galaxy images derived from the Galaxy Zoo citizen science project~\citep{lintott2008galaxy}, specifically the Galaxy Zoo: DECaLS campaign \citep{2022MNRAS.509.3966W}. Images originally have a resolution of $64\times64$ pixels, which are upsampled to $224\times224$ pixels for consistency across our experiments. 

The dataset contains four morphological classes annotated by volunteers: \textit{Smooth Round}, \textit{Smooth Cigar-shaped}, \textit{Edge-on Disk}, and \textit{Unbarred Spiral}. Each galaxy must have received at least 17 votes by human volunteers during either the first or second campaigns to be included in the dataset. Finally, the dataset is balanced such that each morphological class contains 2,500 images each: for a total of 10,000 images in the dataset. While the resulting 25\% anomaly fraction may appear closer to a general classification problem than a typical anomaly detection scenario, this configuration is consistent with the range of class imbalance ratios explored in established anomaly detection benchmarks \citep[e.g.][]{2022arXiv220609426H}, and we treat it as such here.

For evaluation, we treated each morphological type sequentially as anomalies, using the other three classes as normal samples. We started experiments with a small labelled set of 40 images (10 anomalies, 30 normal). As before, each training step utilised an additional randomly sampled pool of 10,000 unlabelled images for training \texttt{AnomalyMatch}.

\subsection{Galaxy Zoo - The Galaxy Challenge}
We also benchmark \texttt{AnomalyMatch} on the 'Galaxy Zoo - The Galaxy Challenge' dataset. This is a subset of $\sim$65,000 galaxy images extracted from Galaxy Zoo 2~\citep{2013MNRAS.435.2835W} for the 'Galaxy Challenge' Kaggle competition. The dataset consists of images of galaxies where citizen scientists classified the galaxy morphology. One of the classifications they made of importance to us was whether the central source was unusual or 'odd'. 

Each galaxy was seen by at least 40 citizen scientists. Each object is ranked by the percentage of users which classified this image accordingly. This test set is chosen to directly compare to the results from \citet{Lochner2021Astronomaly}, who tested the effectiveness of \texttt{Astronomaly} at identifying 'odd' galaxies in this dataset. We follow their approach and define an anomalous galaxy as one for which 90\% of the volunteers classified it as 'odd'. In their work, they obtain a sample of 924 anomalies by applying this cut, and then train their algorithm on 200 of these examples and remove them from the remaining test set. Therefore, the number of anomalies they are aiming to recover is 724. 

\section{Results}\label{results_sect}
\subsection{miniImageNet}\label{results:miniImageNet_sect}
We summarise our experimental results on the miniImageNet dataset, focusing initially on the class \textit{Hourglass} as a representative example. Results for the other evaluated anomaly classes (\textit{Guitar}, \textit{Printer}, \textit{Piano}, and \textit{Orange}) are presented in Table~\ref{tab:miniimagenet_results}.
Given the severe class imbalance (approximately 1\% anomalies), high performance under the efficiency metric is critical to minimise manual inspection by domain experts.

\begin{figure}
    \centering
    \includegraphics[width=\columnwidth]{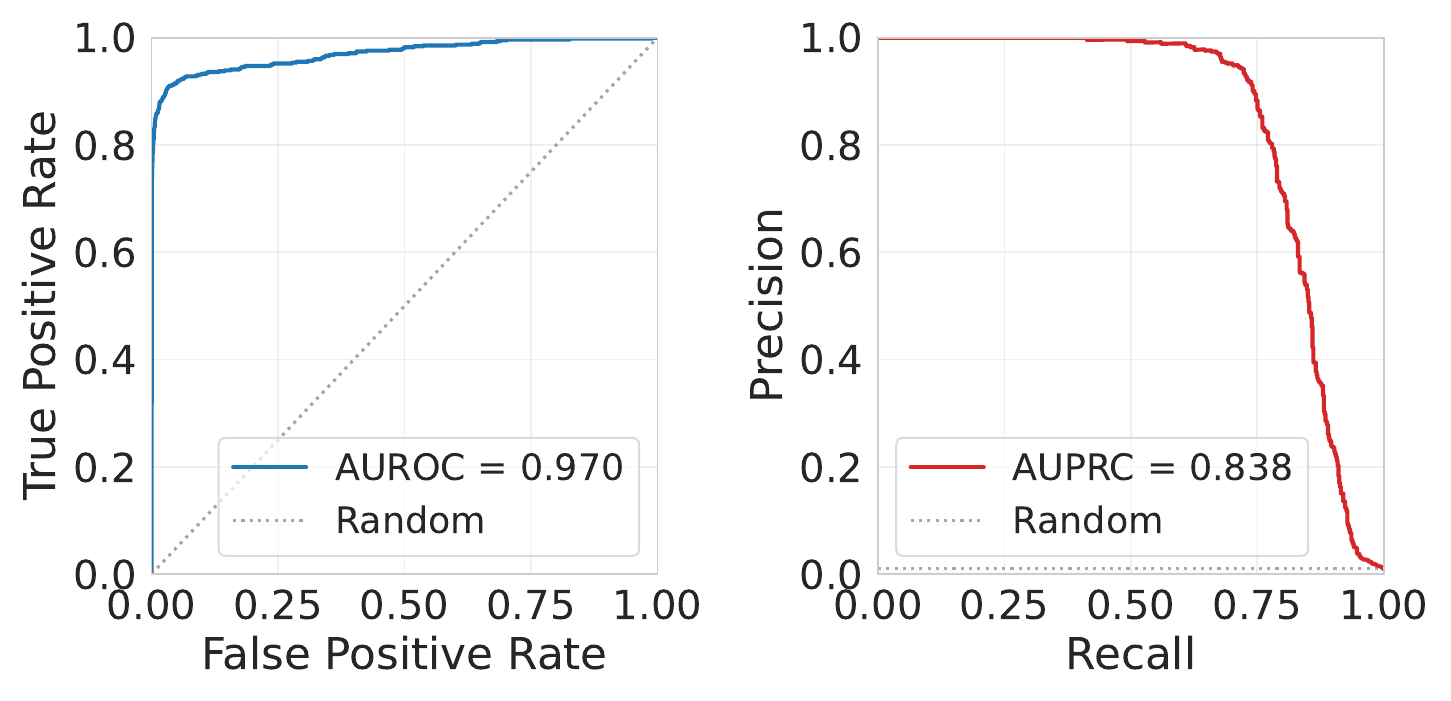}
    \caption{ROC (left) and PR (right) curves after three active learning cycles for the miniImageNet \textit{Hourglass} anomaly class starting from five labelled anomalies, adding ten after each cycle. The model achieves an AUROC of 0.96 and an AUPRC of 0.80, highlighting robust anomaly detection capability under severe class imbalance.}
    \label{fig:roc_prc_hourglass}
\end{figure}

Figure~\ref{fig:roc_prc_hourglass} shows the Receiver Operating Characteristic (ROC) curve -- plotting the TPR against the FPR -- and the Precision-Recall (PR) curve for the \textit{Hourglass} anomaly class after the final active learning cycle. Our approach achieved an AUROC of $0.96$ and AUPRC of $0.83$, indicating strong separation between normal and anomalous samples.

\begin{figure*}
    \centering
    \subfigure[Anomaly Detection Efficiency progression across cycles: efficiency improves quickly, with strong performance already after the first cycle. The red dashed line represents perfect detection performance.]{\includegraphics[width=0.48\textwidth]{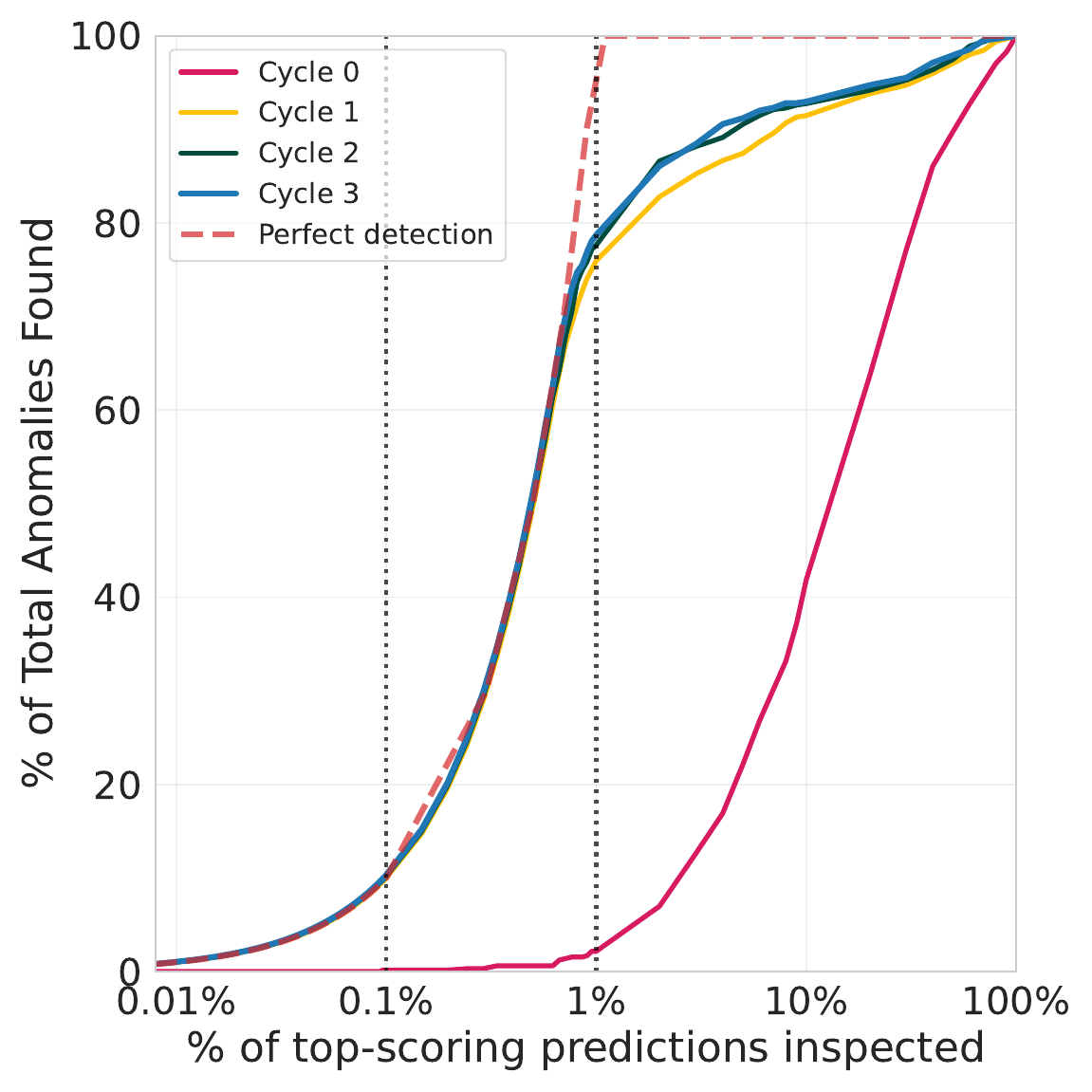}}
    \hspace{0.5cm}
    \subfigure[Final performance after three cycles: 76.4\% of all anomalies are found within the top 1\% of the ranked predictions, with 10.3\% already recovered in the top 0.1\%.]{\includegraphics[width=0.48\textwidth]{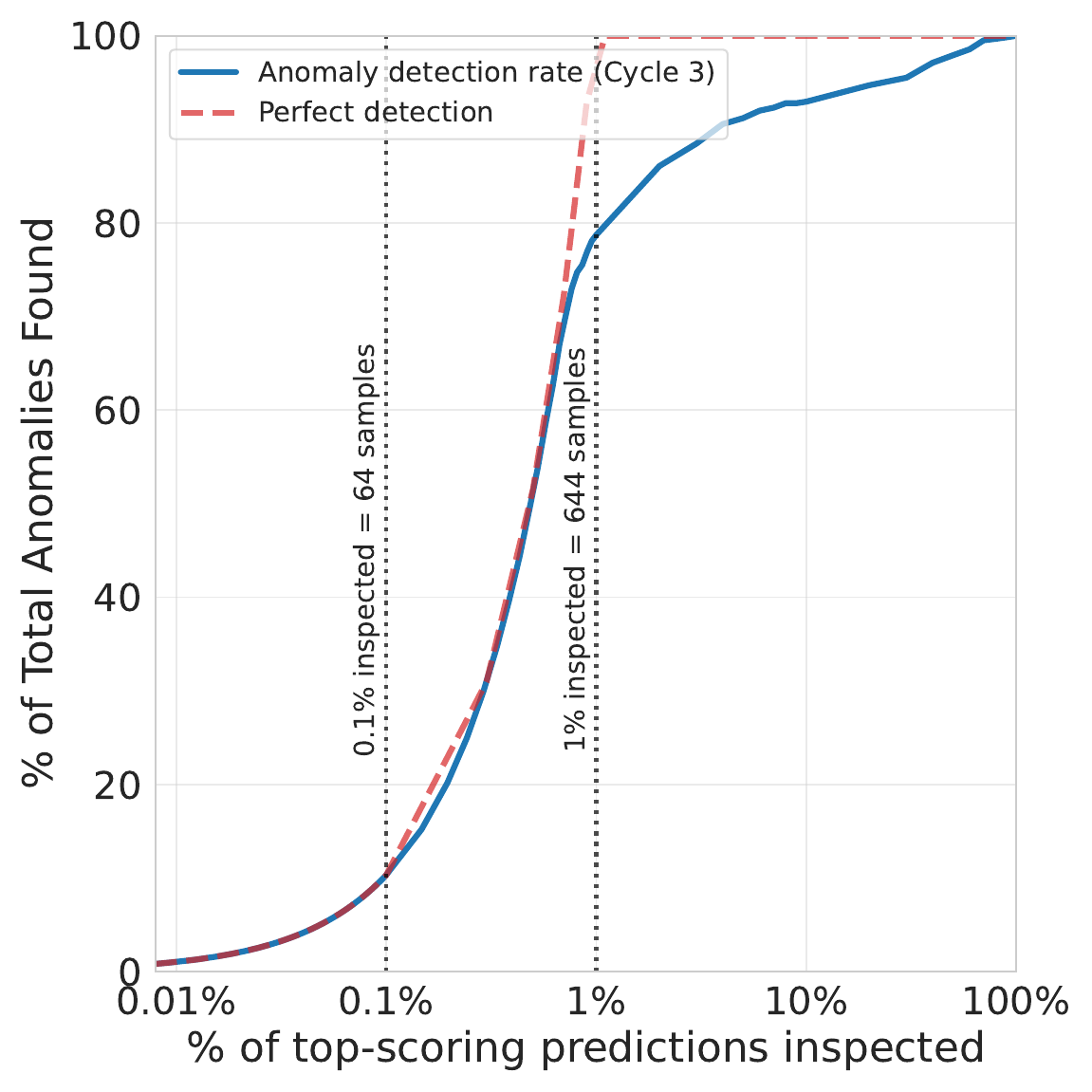}}
    \caption{Anomaly Detection Efficiency curves for the \textit{Hourglass} class in miniImageNet. Each point on the x-axis represents a percentage of the ranked predictions inspected, sorted by descending anomaly score (i.e., the most anomalous images are checked first). The y-axis shows the percentage of true anomalies recovered within the inspected data subset. At 0.1\%, the total number of possible detectable anomalies is $\sim64$, while at 1\% the number of detectable anomalies is all 625 remaining in the class.  Active learning started with five labelled anomalies and adds ten samples after each cycle.}
    \label{fig:miniimagenet_hourglass_combined}
\end{figure*}

Figure~\ref{fig:miniimagenet_hourglass_combined} shows the Anomaly Detection Efficiency curve after three cycles for the \textit{Hourglass} anomaly class. The method identifies 74.7\% of anomalies within the top 1\% highest-scoring images, demonstrating high capability to retrieve the specified anomaly for the user. The top-scoring 0.1\% data samples are all correctly identified anomalies. Successive active learning only marginally increase performance. A likely reason is that learning gains from the initial sampling data, which remain the majority of labelled data throughout, are already achieved within the first cycle.

Table~\ref{tab:miniimagenet_results} summarises performance metrics across all tested anomaly classes. The model consistently achieves high AUROC and AUPRC values, identifying approximately three-quarters of anomalies in the top 1\% scoring images, and attaining precision between 73\% and 81\%. Average precisions for the top-scoring 0.1\% of the data were 100\%. This indicates strong practical utility for human reviewers who would want to quickly identify additional anomalies.

\begin{table*}
\caption{Evaluation on unlabelled data for miniImageNet with different classes used as anomaly after three cycles with active learning. At 100\% precision no more anomalies could have been found in the top-scoring  0.1\% or 1\% of the dataset, respectively. Training started from five labelled anomalies, adding ten after each cycle. We include the number of anomalies detected in the top scoring 100 and 1,000 sources. Note, the total number of anomalies that can be identified is 625 per class.}
\label{tab:miniimagenet_results}
\centering
\begin{tabular}{ccccccc}
\hline
Anomaly Class & AUROC & AUPRC & \thead{Precision in \\ top-scoring 0.1\% [\%]} & \thead{Precision in \\ top-scoring  1\% [\%]} & \thead{Anomalies in \\ Top 100} & \thead{Anomalies in \\ Top 1000} \\
\hline
Guitar & 0.95 & 0.75 & 100.00 & 68.63 & 100.00 & 474.00 \\
Hourglass & 0.97 & 0.84 & 100.00 & 76.40 & 100.00 & 531.00  \\
Printer & 0.97 & 0.85 & 100.00 & 79.35 & 100.00 & 537.00 \\
Piano & 0.97 & 0.83 & 100.00 & 78.26 &  100.00 & 527.00 \\
Orange & 0.95 & 0.82 & 100.00 & 76.55 & 100.00 & 517.00 \\
\hline
Mean & 0.96 $\pm$ 0.01 & 0.82 $\pm$ 0.04 & 100.00 $\pm$ 0.00 & 75.84 $\pm$ 4.21 & 100.00 $\pm$ 0.00 & 517.20 $\pm$ 22.56 \\
\hline
\end{tabular}
\end{table*}

In summary, these results demonstrate the effectiveness of our approach, highlighting its practical value by consistently ranking most anomalies highly and significantly reducing the labelling workload required from human experts for enabling such a result by starting from only five labelled samples.

\subsection{GalaxyMNIST}
\begin{figure}
    \centering
    \includegraphics[width=\columnwidth]{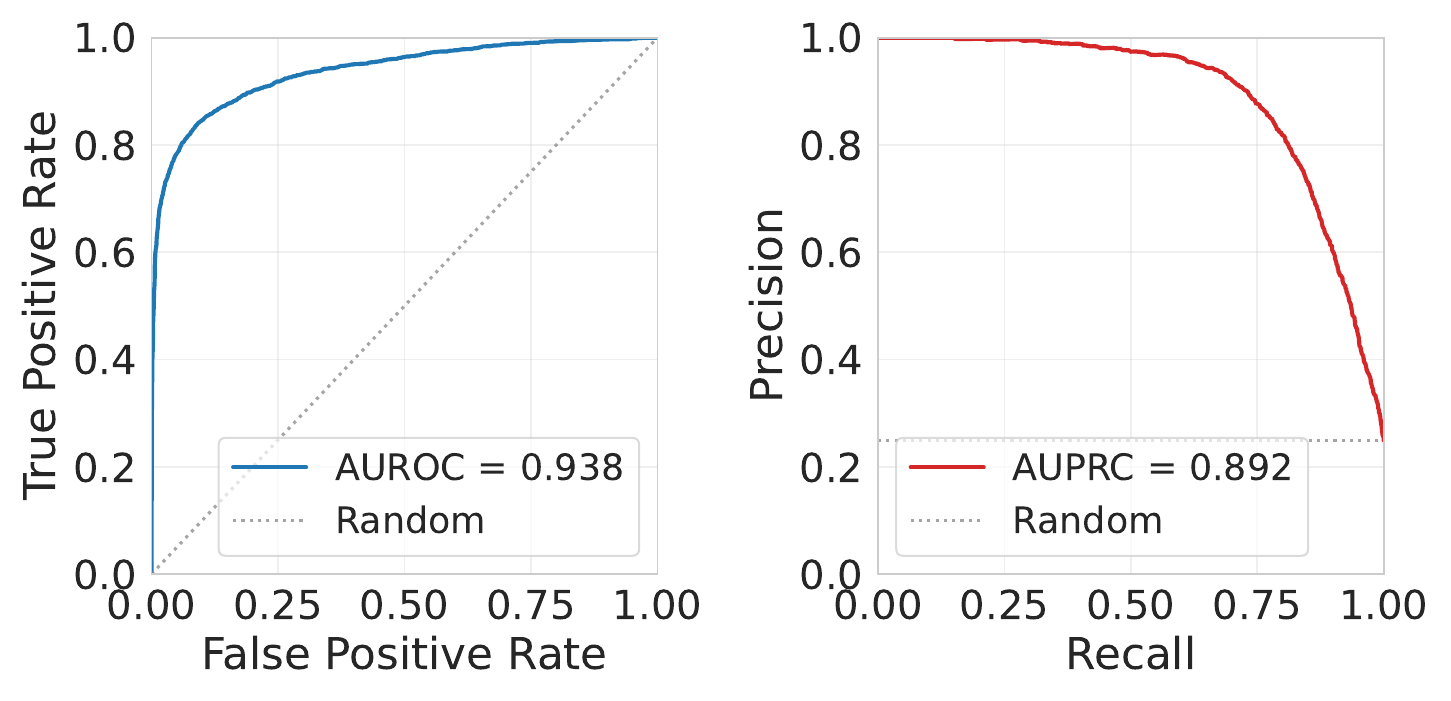}
    \caption{ROC (left) and PR (right) curves for the \textit{Unbarred Spiral} anomaly class from GalaxyMNIST. High AUROC and AUPRC indicate effective anomaly identification despite the inherent morphological similarities across galaxy classes. Training started from ten labelled anomalies, adding ten after each cycle. }
    \label{fig:galaxymnist_roc_prc}
\end{figure}

Figure~\ref{fig:galaxymnist_roc_prc} illustrates our results for the GalaxyMNIST dataset using the \textit{Unbarred Spiral} class as representative anomalies. ROC and PR curves for this class confirm the robustness of the model. Despite GalaxyMNIST's classes being visually similar and challenging to distinguish even for human observers, the model achieves excellent performance metrics (AUROC: 0.89, AUPRC: 0.77), significantly above random chance.

\begin{figure*}
    \centering
    \subfigure[Active learning efficiency across cycles: progressive improvements are visible over three cycles, with more anomalies recovered early in the ranked predictions.]{\includegraphics[width=0.48\textwidth]{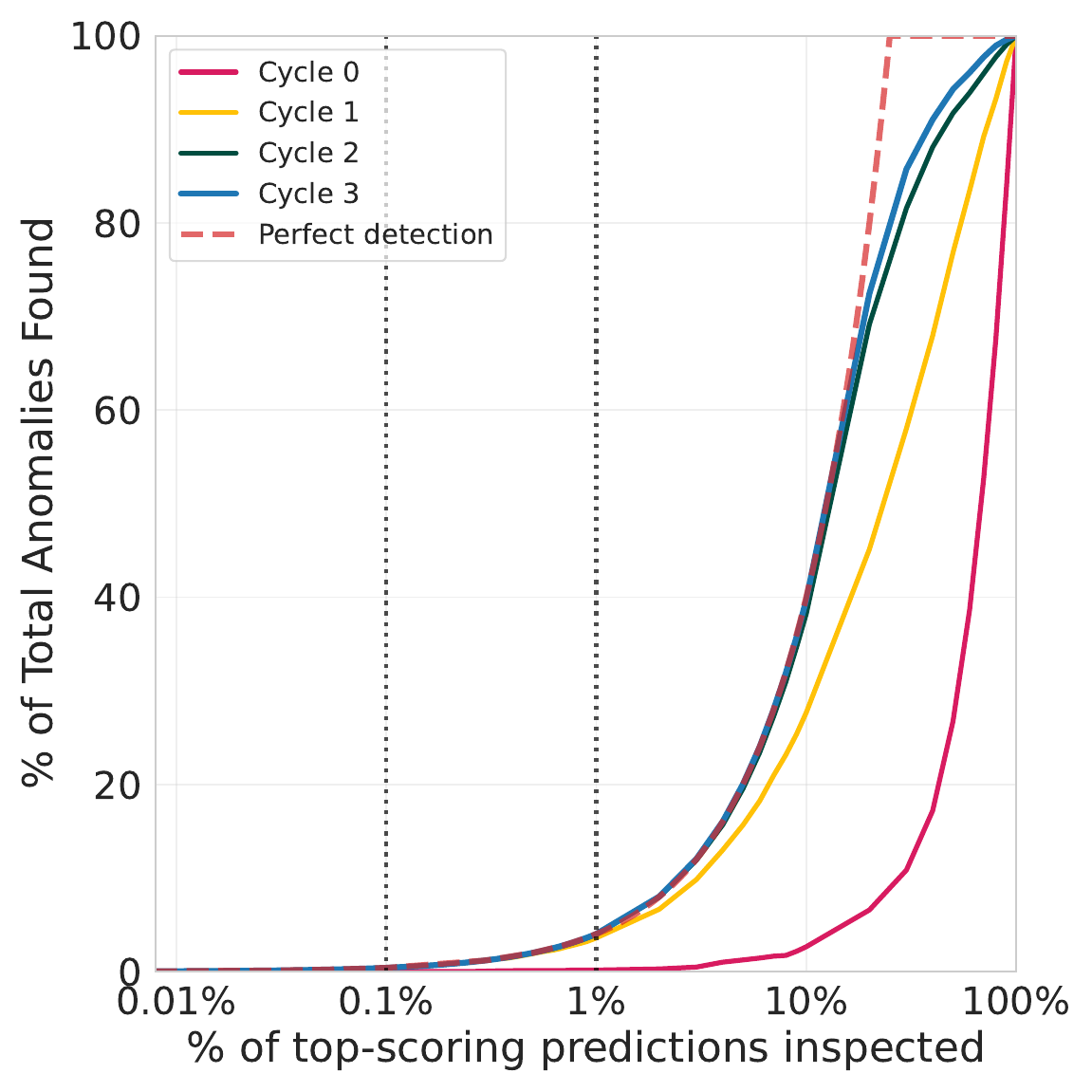}}
    \hspace{0.5cm}
    \subfigure[Final performance after three cycles: within the top 1\% of the ranked predictions, only anomalies are found.]{\includegraphics[width=0.48\textwidth]{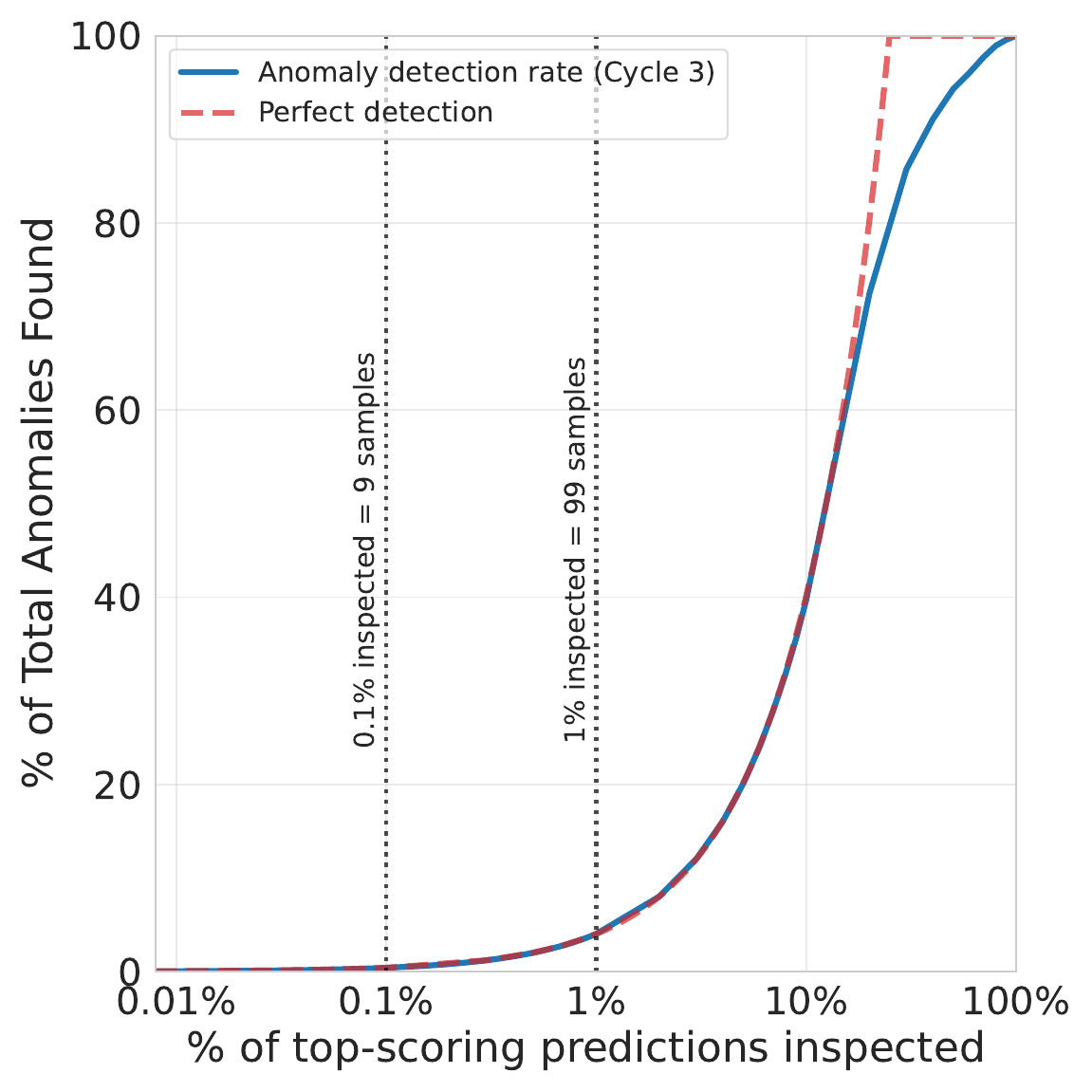}}
    \caption{Anomaly Detection Efficiency curves for the \textit{Unbarred Spiral} class in GalaxyMNIST (25\% of the dataset). The x-axis shows the percentage of the data inspected, sorted by descending predicted anomaly score, while the y-axis reflects the cumulative percentage of true anomalies recovered. At 0.1\%, the total number of detectable anomalies is $\sim9$ while at 1\% it is 99.  The model is trained using ten labelled anomalies initially, with ten more added after each active learning cycle.}
    \label{fig:galaxymnist_unbarred_spiral_combined}
\end{figure*}

Figure~\ref{fig:galaxymnist_unbarred_spiral_combined} a) shows how Anomaly Detection Efficiency evolves across cycles. A clear improvement of the model is visible after each active learning step. The final Anomaly Detection Efficiency curve in Figure~\ref{fig:galaxymnist_unbarred_spiral_combined} b) demonstrates strong final performance, particularly at lower inspection percentages. Remarkably, within the top 0.1\% highest-scoring samples, 100\% precision was achieved, thus successfully identifying anomalies immediately within a very small subset of the data. Successive active learning continue to increase performance in contrast to the experiments on miniImageNet. One potential reason may lie in the overall greater number of anomalous samples in the dataset, leading to a potentially higher heterogeneity of anomalies.

Table~\ref{tab:galaxymnist_results} summarises results across all GalaxyMNIST anomaly classes. 
Overall, anomaly detection performance on GalaxyMNIST varied between classes, reflecting visual and morphological similarities inherent in the dataset. Notably, three out of four classes achieved perfect anomaly detection precision within the top 0.1\% of highest-scoring samples, demonstrating excellent efficiency in scenarios where quick anomaly identification from large data pools is required. Conversely, classes like \textit{Smooth Cigar-shaped} and \textit{Edge-on Disk} led to lower AUROC and AUPRC values, highlighting the inherent difficulty and potential ambiguity in visually separating these morphologies.

\begin{table*}
\caption{Evaluation on unlabelled data for GalaxyMNIST with different classes used as anomaly after three cycles with active learning. At 100\% precision no more anomalies could have been found in the top-scoring 0.1\% or 1\% of the dataset, respectively. We include the number of anomalies found in the top scoring 100 and 1,000 images. Training started from ten labelled anomalies, adding ten after each cycle.}
\label{tab:galaxymnist_results}
\centering
\begin{tabular}{ccccccc}
\hline
Anomaly Class & AUROC & AUPRC & \thead{Precision in \\ top-scoring 0.1\% [\%]} & \thead{Precision in \\ top-scoring  1\% [\%]} & \thead{Anomalies in \\ Top 100} & \thead{Anomalies in \\ Top 1000} \\ \\
\hline
Smooth Round & 0.98 & 0.95 & 100.00 & 100.00 & 100.00 & 998.00 \\
Smooth Cigar-shaped & 0.74 & 0.51 & 88.89 & 78.79 & 79.00 & 643.00 \\
Edge-on Disk & 0.89 & 0.74 & 88.89 & 96.97 & 97.00 & 851.00 \\
Unbarred Spiral & 0.94 & 0.89 & 100.00 & 100.00 & 100.00 & 988.00 \\
\hline
Mean & 0.89 $\pm$ 0.10 & 0.77 $\pm$ 0.20 & 94.44 $\pm$ 6.41 & 93.94 $\pm$ 10.20 & 94.00 $\pm$ 8.84 & 870.00 $\pm$ 143.35 \\

\hline
\end{tabular}
\end{table*}

\subsection{Galaxy Zoo - The Galaxy Challenge}
In our case for training, we randomly sampled (with a fixed seed) three distinct subsets from the overall Galaxy Challenge dataset:
\begin{itemize}
    \item 100 normal and 100 anomalous objects;
    \item 200 normal and 200 anomalous objects;
    \item 394 normal and 6 anomalous objects.
\end{itemize}
After each training cycle, we correct the top 10 highest-scoring mislabelled normal and confirm the top 10 highest-scoring anomalous objects. To achieve robust measurements of our performance, we run these different configurations with 1 well used seed and 8 random seeds\footnote{The following seeds were used, and no further seeds were run: 42, 76032, 730, 83209, 13798, 4538, 5923, 99271, 3762} and average their results. This provides us with better estimates of performance and the variation in our method over different selections of training data, and different selections of active learning sets that are informing the model.

\begin{figure}
    \centering
    \includegraphics[width=0.9\linewidth]{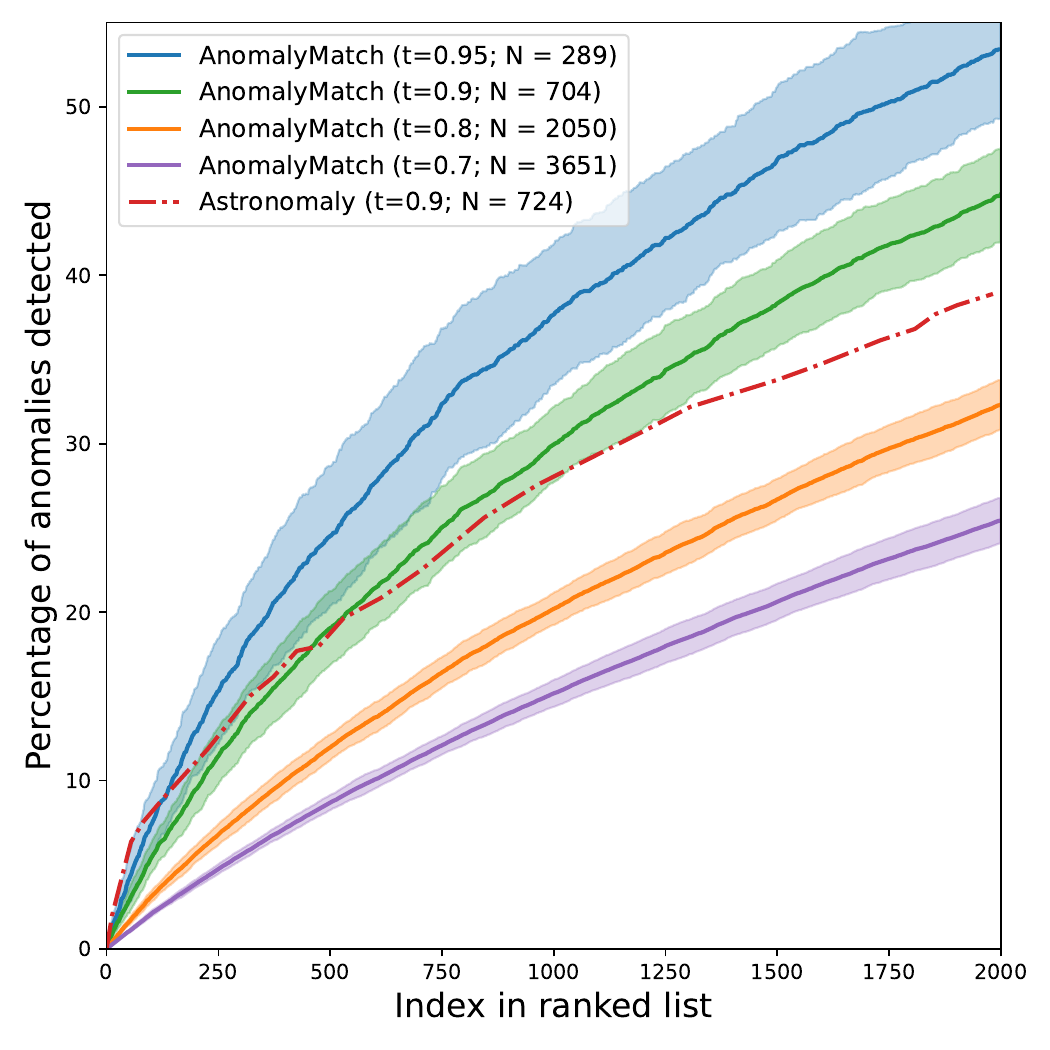}
    \caption{Direct comparison of the percentage of anomalies recovered by \texttt{AnomalyMatch} and \texttt{Astronomaly}. We show the results of an initial training set of 200 normal and 200 anomalous label images and adding 20 labels per active learning cycle in three training cycles, for a total of 440 labels. For \texttt{AnomalyMatch}, we plot the mean curves over 9 random seeds, with shaded regions indicating the standard deviation. The curve for \texttt{Astronomaly} (dashed red) is taken from Figure 5a of \citet{Lochner2021Astronomaly}. We find that the performance of \texttt{Astronomaly} and \texttt{AnomalyMatch} is comparable, although \texttt{AnomalyMatch} performs slightly worse at initial ranks but then starts to outperform \texttt{Astronomaly}. Varying the voter fraction threshold leads to a differing performance of \texttt{AnomalyMatch}.}
    \label{fig:200-200_astronomaly_comparison}
\end{figure}

Figure~\ref{fig:200-200_astronomaly_comparison} shows the percentage of anomalies identified in the top N-ranked images based on the score given by \texttt{AnomalyMatch} using a training set of 200 nominal and 200 anomaly images (bringing the total to 440 after two rounds of active learning adding 10 nominal and 10 anomaly images each). The rank here is a simple ordering of the dataset from most anomalous to least anomalous by score, where objects with identical score will be randomised in their ordering. As stated previously, we have a soft truth label of what is and is not an anomaly by taking the 90\% voter fraction cut used in \citet{Lochner2021Astronomaly}. This corresponds to searching for 704 anomalies, once the training set is accounted for. We find that, overall, our performance is similar to that of \texttt{Astronomaly}. At high ranks (low index), our performance is below that of \texttt{Astronomaly} while at lower rank (increasing index) \texttt{AnomalyMatch} begins to outperform it. However, both methods are below a perfect 1:1 detection. 

To investigate the cause of this low performance, we vary the definition of an anomaly by applying different voter fraction (GZ Score; $t$) cutoffs. We find that the performance is sensitive to the chosen cutoff. Altering this voter cutoff affects our performance as the number of anomalies we are attempting to identify changes. For instance, with a voter cutoff of 0.8 ($N_{\text{anomalies}} = 2,050$) we detect 662 anomalies ($\sim32\%$) in the top 2,000 scored images. Similarly, for 0.7 ($N_{\text{anomalies}} = 3,651$) we detect 928 anomalies ($\sim25\%$) in the top 2,000 scores. Making the cutoff more stringent, to 0.95, reduces the number to search for to $N_{\text{anomalies}} = 289$, where we detect 154 ($\sim53\%$) in the top 2,000 ranked images. This shows the volatility in the definition of an anomaly in this dataset. Furthermore, we cannot directly compare this to  \texttt{Astronomaly}, as they do not investigate the variance in their model performance nor the effect of changing the voter cutoff.

\begin{figure}
    \centering
    \includegraphics[width=1.0\linewidth]{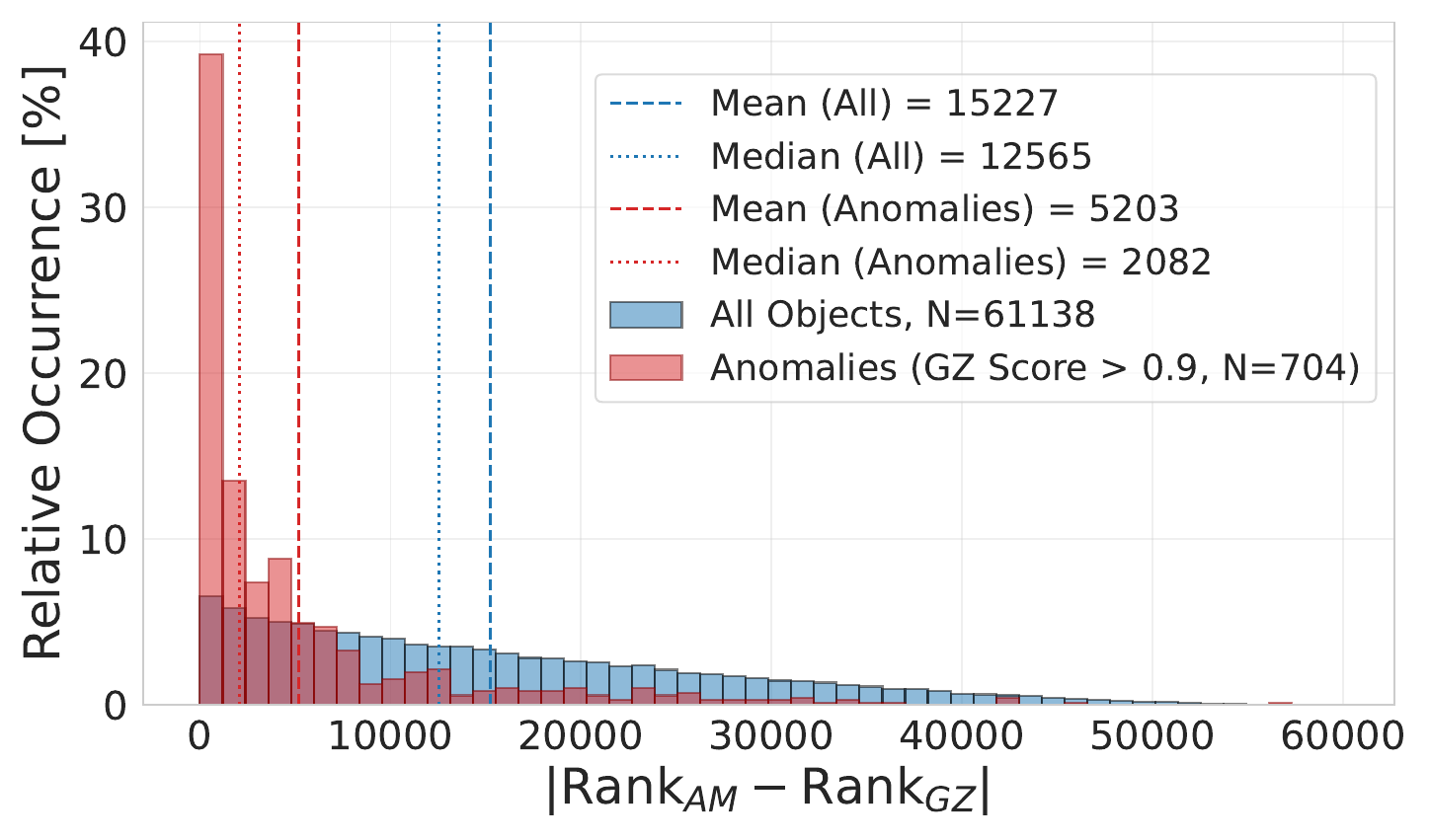}
    \caption{Comparison of the ranks based on GZ scores and scores from \texttt{AnomalyMatch} for fixed Seed 4538 after Cycle 3 with the 200 - 200 balanced labelled set. It is evident that the ranks assigned by \texttt{AnomalyMatch} are relatively close to the user ranks in the case of anomalies, while uninteresting objects have a much broader distribution.}
    \label{fig:200-200_rank_comparison}
\end{figure}

To explore the agreement or disagreement between the volunteers and \texttt{AnomalyMatch}, we investigate differences in ranking given by each. We subtract the citizen scientist rank from the rank found by \texttt{AnomalyMatch}. If the Galaxy Zoo rank and \texttt{AnomalyMatch} rank are in agreement, this difference is close to zero. If the citizen scientists and \texttt{AnomalyMatch} disagree, this rank difference is large. Figure~\ref{fig:200-200_rank_comparison} shows the distribution for the entire sample (blue) and only the objects defined as anomalies (red). We find that the volunteers and \texttt{AnomalyMatch} are in close agreement (rank difference $\leq 1000$) for 273 ($\sim$38\%) of anomalies. 

\begin{figure*}
    \centering
    \includegraphics[width=1.0\textwidth]{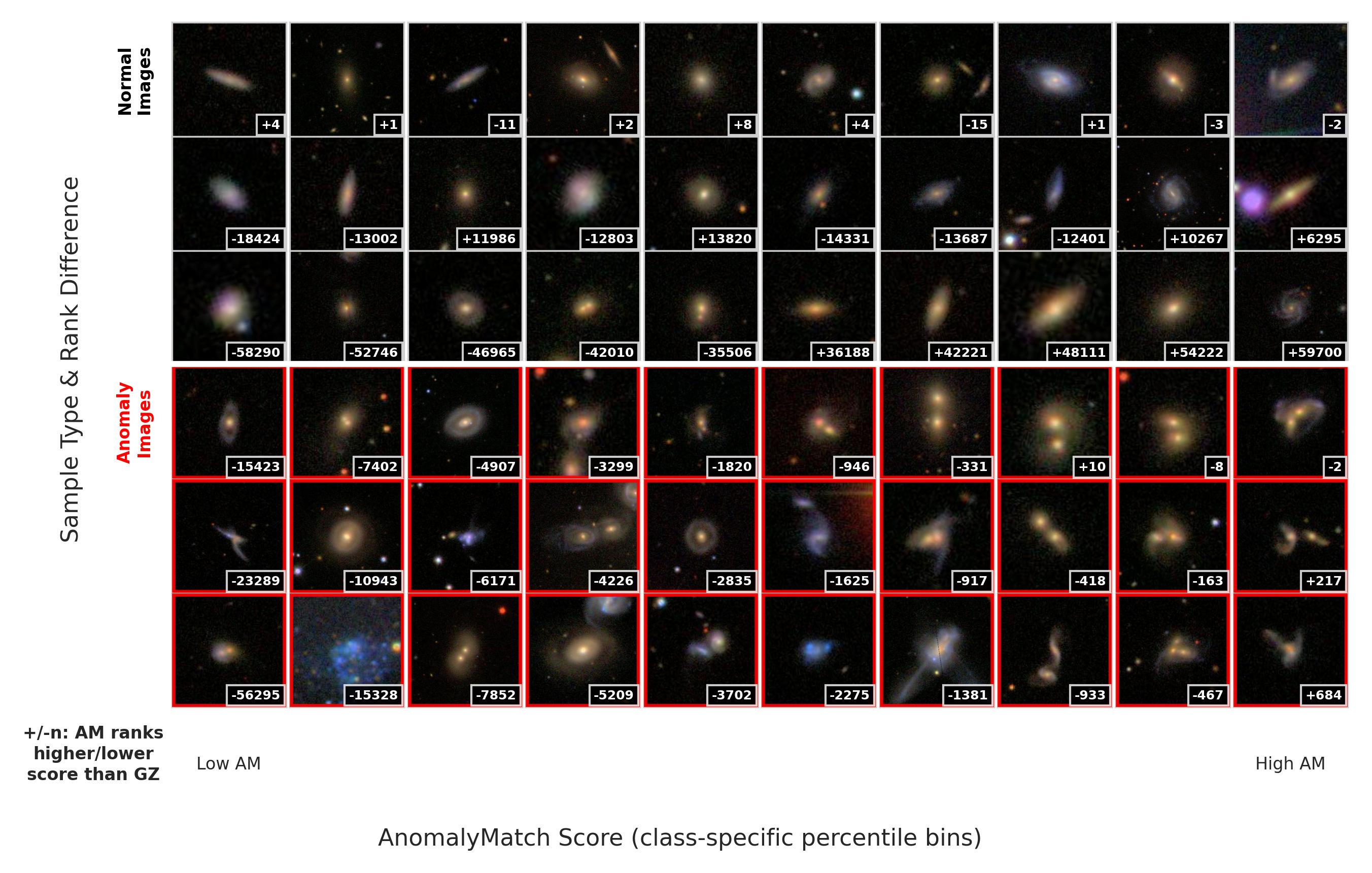}
    \caption{Comparison of the GZ ranks and the \texttt{AnomalyMatch} scores with fixed Seed 4538 after Cycle 3 for the 200 - 200 balanced labelled set. Each column represents a percentile of AnomalyMatch scores (class based) and displays three samples (closest -, median - and furthest absolute rank deviation) for normal and anomaly images. Positive rank differences indicate that \texttt{AnomalyMatch} gave the object a higher score.
    In the anomalies it can be seen, that objects showing strong tidal features were typically ranked/scored high and closer to the GZ rank, while overlapping galaxies or galaxies with 2 cores were ranked much lower. Of the normal images noisy galaxies showing colour gradients were often ranked much higher than in GZ 2. In a real world use case a user would recognise such clear mislabelling and amend the dataset accordingly. To ensure the central object is interpretable in this figure, we have cropped each image by 40 pixels on each side. This allows the reader to more clearly see the central object.}
    \label{fig:200-200galaxyzoo_grid_comparison}
\end{figure*}

Figure~\ref{fig:200-200galaxyzoo_grid_comparison} shows a grid of example sources ordered by agreement between \texttt{AnomalyMatch} scores and GZ Scores. Objects are sorted from left to right according to the anomaly score assigned by \texttt{AnomalyMatch}. Vertically, the objects are split into normal (top) and anomalies (bottom). The three rows in each group correspond (from top to bottom) to minimal, median and highest rank disagreement. The difference between the volunteer and \texttt{AnomalyMatch} ranks are shown in the inset of each image. For the anomalies (red borders) we find that the disagreement can, for example, come from dual core objects, as well as mergers showing limited tidal features. Objects scored highly by \texttt{AnomalyMatch}, in general, are often those with clear tidal features and ongoing interactions.

There is a significant difference in volunteer votes and \texttt{AnomalyMatch} scores for images with colour errors, images containing artifacts and ring galaxies, to which \texttt{AnomalyMatch} assigned low scores. These were likely not represented in our training sample and not included while conducting active learning. The bottom row of the normal images in Figure \ref{fig:200-200galaxyzoo_grid_comparison} shows that \texttt{AnomalyMatch} disagreed with volunteers by ranking as anomalous noisy images or ones with small and close, possibly interacting, companions and weak but possible tidal features. Meanwhile, anomaly images of overlapping sources were typically ranked much lower (bottom left) than by users. The disagreement in the normal class likely stems from the difficulty for the volunteers to visually identify merger signatures that are more easily found by \texttt{AnomalyMatch}.

\begin{figure}
    \centering
    \includegraphics[width=1.0\linewidth]{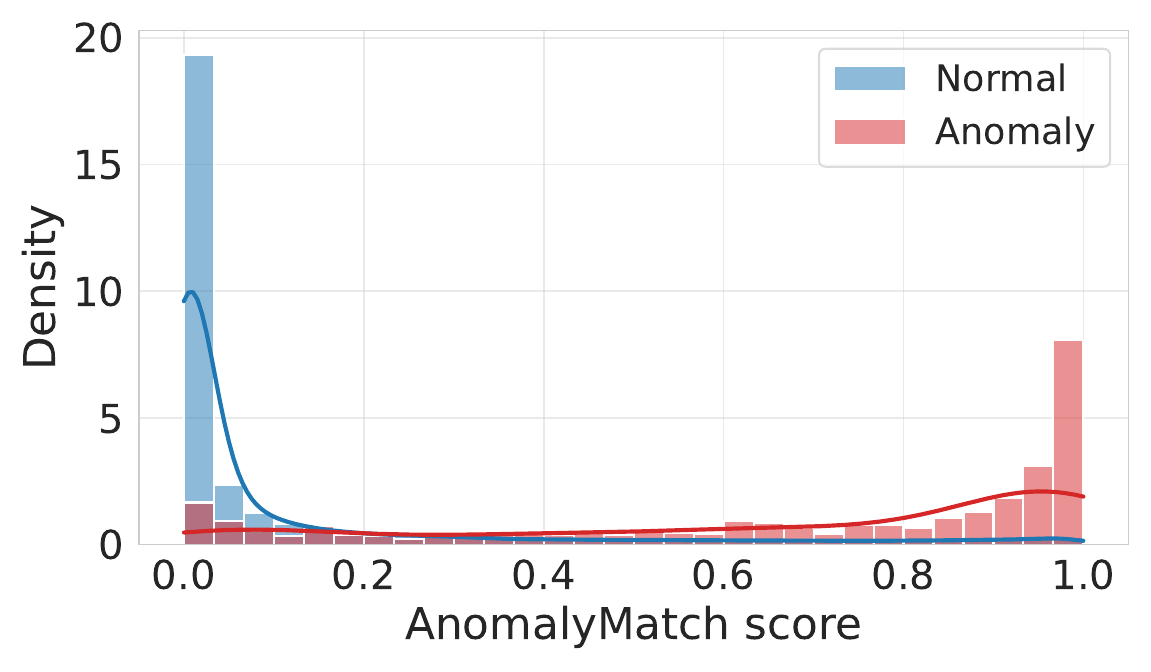}
    \caption{Histogram of the scores computed by \texttt{AnomalyMatch} with a specific seed after cycle 3 with the 200 - 200 balanced labelled set. A very clear separation between anomalies and the normal images is present.}
    \label{fig:200-200_score_histogram}
\end{figure}

Figure \ref{fig:200-200_score_histogram} shows the distribution of scores given by \texttt{AnomalyMatch} to normal and anomalous images as ranked by the volunteers. The score distribution for normal images (blue) is highly skewed towards low scores. In contrast, the distribution for anomalies (red) is comparatively more bimodal. While most anomalies receive high scores, there is also a small overdensity at low scores. Examples of anomalies that fall into this overdensity - i.e. anomalies `missed' by \texttt{AnomalyMatch} - are the dual-core galaxies in the bottom left of Figure \ref{fig:200-200galaxyzoo_grid_comparison}. This also reinforces that during active learning a user should also consider what is being misclassified by the model to correct its behaviour.

Table~\ref{tab:galaxyzoo_results} summarises our performance metrics across different breakdowns of labelled vs unlabelled data in training. The best performance is achieved by the evenly split model with 200-200 anomaly-normal labels, with top-scoring 0.1\% precision of $(40 \pm 10) \%$. For the more likely real-world setup of 6-394, we measure mean precisions of $(31.93 \pm 9.84)$\% and $(19.27\pm 3.69)$\% in the top-scoring 0.1\% and 1\% of anomalies observed. We also add here the number of anomalies detected in the top 100 and top 1000 scored images, averaged across each seed. 

\begin{table*}
\caption{Evaluation on unlabelled data for the Galaxy Zoo 2 dataset with different initial configurations, varying in the number of labels and the proportion of anomalies. All results are averaged over the 9 random seeds.}
\label{tab:galaxyzoo_results}
\centering
\begin{tabular}{ccccccc}
\hline
\thead{Labels \\ {[Total (Anomalies / Normal)]}} & AUROC & AUPRC & \thead{Precision in \\ top-scoring 0.1\% [\%]} & \thead{Precision in \\ top-scoring  1\% [\%]} & \thead{Anomalies in \\ Top 100} & \thead{Anomalies in \\ Top 1000} \\
\hline
200 (100 / 100) & 0.87 $\pm$ 0.02 & 0.12 $\pm$ 0.01 & 32.60 & 20.41 & 30.22 & 171.22 \\
400 (200 / 200) & 0.91 $\pm$ 0.01 & 0.17 $\pm$ 0.02 & 39.52 & 25.00 & 39.22 & 209.78   \\
400 (6 / 394) & 0.71 $\pm$ 0.04 & 0.06 $\pm$ 0.03 & 23.67 & 12.40 & 21.89 &  104.89 \\
\hline
Mean & 0.83 $\pm$ 0.02 & 0.12 $\pm$ 0.01 & 31.93 $\pm$ 9.84 & 19.27 $\pm$ 9.18 & 30.44 $\pm$ 7.08 & 161.89 $\pm$ 43.32 \\
\hline
\end{tabular}
\end{table*}

\subsection{Number of Initial Labels}
We investigated the impact of varying the number of initial labels while maintaining the proportion of anomalies within these labels. Three scenarios were tested: 100 initial labels (one anomaly, 99 normal), 500 initial labels (five anomalies, 495 normal), and 1000 initial labels (10 anomalies, 990 normal). After each active learning cycle, we correspondingly increased the number of actively labelled samples per iteration proportionally to the initial sample size: 10 additional labels per iteration for the smallest set, 20 labels for the intermediate set, and 40 labels for the largest set.

Detailed results for these ablations are summarised in Table~\ref{tab:n_samples_ablation_results}. Interestingly, we observed limited differences in performance metrics across all tested initial labelling conditions. AUROC values varied by less than $0.03$, and Anomaly Detection Efficiency remained consistently high. Notably, all experiments achieved perfect precision at the stringent threshold of inspecting just the top 0.1\% scoring data points, further underscoring the robustness and consistency of our active learning pipeline. This indicates that our approach effectively utilises available labels and does not significantly benefit from substantially larger initial label sets, highlighting its practical efficacy even with minimal labelled data.

\begin{table*}
\caption{Effect of size of initial labelled dataset on anomaly detection performance in miniImageNet. Each configuration starts with a different number of labelled samples and adds different number of labels per cycle. Despite large differences in initial label counts, performance remains robust across metrics, with perfect precision at the top 0.1\% most anomalous samples for all scenarios. Beyond 500 (+20) labels only minimal performance gains are observed.}
\label{tab:n_samples_ablation_results}
\centering
\begin{tabular}{ccccccc}
\hline
\thead{Labels \\ {[Total (Anomalies / Normal)]}} & AUROC & AUPRC & \thead{Precision in \\ top-scoring 0.1\% [\%]} & \thead{Precision in \\ top-scoring  1\% [\%]} & \thead{Anomalies in \\ Top 100} & \thead{Anomalies in \\ Top 1000} \\
\hline
100 (1 / 99) + 10 per iter. & 0.92 & 0.35 & 59.38 & 41.82 & 63.00 & 359.00 \\
500 (5 / 495) + 20 per iter. & 0.95 & 0.76 & 100.00 & 69.88 & 100.00 & 483.00 \\
1000 (10 / 990) + 40 per iter & 0.96 & 0.80 & 100.00 & 71.52 & 100.00 & 476.00 \\
\hline
Mean & 0.94 $\pm$ 0.02 & 0.64 $\pm$ 0.25 & 86.46 $\pm$ 23.45 & 61.07 $\pm$ 16.69 & 87.60 $\pm$ 17.44 & 439.33 $\pm$ 56.88 \\
\hline
\end{tabular}
\end{table*}

\subsection{Training Iterations}
To test for the optimal number of iterations in our setup, we varied the number of \texttt{FixMatch} training iterations per cycle across four settings: 50, 100, 250, and 500. Table~\ref{tab:training_iterations_results} shows the results of these experiments. We observe that performance peaks at around 100 training iterations per cycle. Using more iterations results in slightly worse AUROC and AUPRC, which we attribute to overfitting on the initially provided labelled data. As the active learning loop dynamically updates the training set, excessively long training phases may lead to overfitting on initial samples. 


\begin{table*}
\caption{Impact of \texttt{FixMatch} training duration per active learning step. Performance peaks at shorter training cycles (50–100 iterations), with diminishing returns and overfitting at longer durations. This supports our design choice of brief, iterative updates with rapid model adaptation.}
\label{tab:training_iterations_results}
\centering
\begin{tabular}{ccccccc}
\hline
Iterations & AUROC & AUPRC & \thead{Precision in \\ top-scoring 0.1\% [\%]} & \thead{Precision in \\ top-scoring  1\% [\%]} & \thead{Anomalies in \\ Top 100} & \thead{Anomalies in \\ Top 1000} \\
\hline
50 & 0.98 & 0.81 & 100.00 & 75.31 & 100.00 & 522.00 \\
100 & 0.95 & 0.76 & 100.00 & 69.72 & 100.00 & 486.00 \\
250 & 0.92 & 0.73 & 100.00 & 67.55 & 100.00 & 459.00 \\
500 & 0.93 & 0.73 & 100.00 & 67.70 & 100.00 & 461.00 \\
\hline
Mean & 0.95 $\pm$ 0.03 & 0.76 $\pm$ 0.04 & 100.00 $\pm$ 0.00 & 70.07 $\pm$ 3.63 & 100.00 $\pm$ 0.00 & 482.00 $\pm$ 25.43 \\
\hline
\end{tabular}
\end{table*}

\section{Discussion}\label{discussion_sect}
Our results highlight the practical efficiency and robustness of the \texttt{AnomalyMatch} framework, demonstrating significant potential to substantially reduce the manual review burden for anomaly identification tasks. The method efficiently leverages both computational resources and human labelling effort, achieving strong performance even with limited labelled data. Crucially, our experiments reveal diminishing returns from significantly increasing the number of initial labels. This effect is consistent with earlier findings in semi-supervised methodologies and empirical studies that suggest performance gains saturate quickly beyond a modest label count~\citep{Oliver2018RealisticSSL}. This outcome underscores the method's effective utilisation of limited supervision, suggesting that label sparsity -- a common real-world constraint -- is not a major limitation in practice.

A key finding of our study is the pronounced risk of overfitting when extending training beyond the optimal number of iterations per active learning cycle. Notably, these suggest the ideal balance is around 100 iterations per active learning cycle; longer cycles appear to reduce overall Anomaly Detection Efficiency, highlighting the need for careful tuning and early stopping strategies. Such overfitting risk in semi-supervised learning has been previously observed, especially in low-label regimes without strong regularisation~\citep{arazo2020pseudo}. Given that original \texttt{FixMatch} and its derivatives such as \texttt{MSMatch} typically employ extensive training iterations, our adaptation of shorter, frequent retraining cycles emerges as a beneficial modification tailored to active learning scenarios with heavily imbalanced classes and a simplified binary classification task.

Further, preliminary findings from an accompanying paper focused on the \textit{Hubble Space Telescope} dataset~\citep{HSTPaper} suggest that \texttt{AnomalyMatch} might remain effective with even fewer initial labels, further reducing labelling costs. There, we also utilised a hybrid approach: initially using active learning to identify a sufficient number of anomalies, then transitioning to a fully semi-supervised regime for optimal performance.

\begin{figure*}
    \centering
    \includegraphics[width=0.95\textwidth]{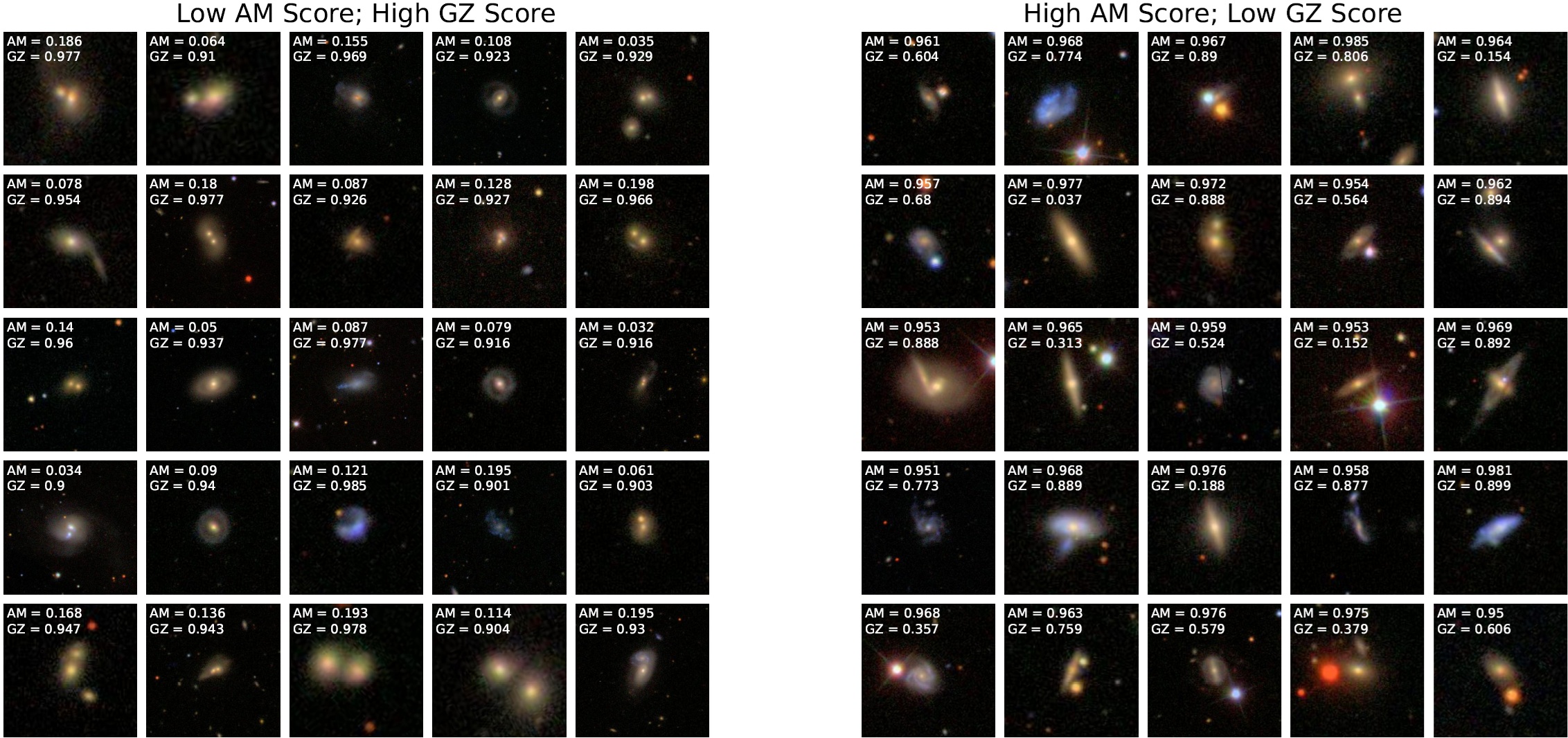}
    \caption{A sample of 50 images given either high score by \texttt{AnomalyMatch} but a low one in Galaxy Zoo 2 or vice versa. \textit{Left}: \texttt{AnomalyMatch} gave a low score ($<0.2$) but the object was classified as an anomaly (Galaxy Zoo score $>0.90$). \textit{Right}: \texttt{AnomalyMatch} gave a high score ($>0.95$) but this was not classified as an anomaly (Galaxy Zoo score $<0.90$). The total number of images satisfying both criteria is 194.}
    \label{fig:incorrect_GZ2_images}
\end{figure*}

While we have demonstrated effectiveness across datasets with drastically different class imbalance characteristics -- extremely skewed (miniImageNet) and moderately balanced (GalaxyMNIST) -- further research is necessary to investigate the lower bounds of label numbers needed for robust anomaly detection performance. Our performance evaluation on the Galaxy Zoo 2 dataset is more complex. To explore potential reasons behind the performance metrics we find, we identify the anomalies that \texttt{AnomalyMatch} scored lowly and the normal images that it scored highly. Figure \ref{fig:incorrect_GZ2_images} shows such images: on the left are images that the volunteers scored $>0.90$ while \texttt{AnomalyMatch} gave a score of $<0.2$ (N$ = 112$), the right shows images \texttt{AnomalyMatch} gave scores $>0.95$ and the volunteers scored $<0.90$ (N$ = 82$). 

The left of Figure \ref{fig:incorrect_GZ2_images} shows anomalies that \texttt{AnomalyMatch} "missed". A key example in this representative set is that of ring galaxies. It is likely that this class was not represented in active learning. In practice, a user could correct these disagreements in the active learning stage and therefore circumvent such misclassifications, improving measured performance.

The right hand side of Figure \ref{fig:incorrect_GZ2_images} shows objects that \texttt{AnomalyMatch} scored $>0.95$, but were not classified as anomalies in the Galaxy Zoo 2 dataset. We find many `odd' galaxies, such as a polar ring galaxy (last in third row) as well as many overlapping galaxies that could be of interest. There are also examples of galaxies overlapped with point sources or containing speckles.  It is important to note the Galaxy Zoo 2 scores given to these objects. Many objects have a volunteer vote fraction close to 0.90, our chosen threshold for separating nominal from anomalous objects. These edge cases may therefore be real anomalies that were classified as nominal, reducing our performance. While these examples are drawn from our testing set, it is likely that similar misclassifications exist in the Galaxy Zoo 2 dataset used for training. Individual misclassifications in the training set can affect the model's ability to correctly identify edge cases.

\subsection{Limitations}
Despite strong results, our study acknowledges several important limitations that warrant further exploration. Addressing these limitations in future work will further strengthen the practical applicability of our approach, enabling its robust deployment in diverse, realistic scientific and industrial contexts.

\textbf{Idealised Active Learning Assumptions.}
Our experiments assumed an idealised active learning environment where human annotations are consistently accurate. In practical deployments, labelling errors and inconsistencies are likely, particularly in ambiguous cases. Additionally, our experiments assumed balanced incremental labelling (equal numbers of anomalies and normal cases per active learning cycle), which may differ significantly in real-world usage where normal labels are generally easier and cheaper to obtain than anomalies. Therefore, we would expect errors and a loss of accuracy to be introduced in a real-world scenario conducting active learning.

\textbf{Realism of Datasets in Anomaly Detection.}
It is important to consider how well each of the datasets used in this work approximates a real-world anomaly search. The Galaxy Zoo 2: The Galaxy Challenge dataset is the closest analogue to a genuine anomaly detection problem: it consists of soft labels derived from volunteer votes, the anomalous class represents a small fraction of the overall dataset, and there is inherent ambiguity in what constitutes an anomaly - as reflected in our finding that the choice of voter fraction threshold significantly affects performance. The GalaxyMNIST dataset, by contrast, is better characterised as a morphological classification benchmark than an anomaly detection problem, with each class comprising 25\% of the total dataset. Additionally, class confusion between the \textit{Edge-on Disk} and \textit{Smooth Cigar-shaped} categories introduces further complexity, which is reflected in our results. The miniImageNet dataset most closely approximates a realistic anomaly detection scenario, with the anomaly class comprising 1\% of the total dataset. Crucially, both miniImageNet and GalaxyMNIST use hard, unambiguous labels with clear class boundaries, and we would therefore expect higher performance on these datasets than on the Galaxy Zoo 2 dataset, where label ambiguity is an inherent feature of the volunteer classification process.

\textbf{Pretraining Bias.}
Performance on the miniImageNet dataset likely benefits from the EfficientNet backbone's ImageNet pretraining. Such pretrained representations inherently capture generic features beneficial for image classification, possibly inflating performance metrics. Future studies should assess performance sensitivity when employing randomly initialised neural networks or representations pretrained on domain-specific data.

\textbf{Limited Exploration of Class Imbalance Scenarios.}
While we investigated datasets at two extremes of class imbalance (1\% vs. 25\% anomalies), real-world anomaly scenarios may present even more extreme imbalance, significantly affecting performance. Additional research is needed to characterise model behaviour across a broader spectrum of imbalance ratios.

\textbf{Anomaly Homogeneity Assumption.}
We simplified the anomaly detection task by treating all anomalies as belonging to a single implicit category distinct from normal instances. This assumption would limit applicability in scenarios with highly heterogeneous anomalies, each requiring distinct learned representations. Incorporating methods explicitly capable of handling diverse anomaly categories could enhance generalisation.

\textbf{\texttt{AnomalyMatch} Generalisability}. We have found that \texttt{AnomalyMatch} is able to generalise well beyond the initially targeted anomaly class. As demonstrated in \citet{HSTPaper}, beginning from a training set of only three edge-on protoplanetary disk examples, the active learning loop naturally surfaced a diverse range of morphologically unusual objects, ultimately enabling the identification of 17 distinct anomaly categories across the \textit{Hubble} Legacy Archive. This suggests that, while \texttt{AnomalyMatch} is designed for targeted anomaly discovery, broader distributional anomaly detection is achievable as a by-product of the active learning process.

\textbf{Impact of Labelling Errors.}
Our study does not examine the robustness of \texttt{AnomalyMatch} to labelling inaccuracies, an important consideration given the potential for human annotation errors during active learning. Exploring training strategies that explicitly handle or mitigate labelling noise is necessary for real-world robustness.

\textbf{Variation in Model Seeds}
In this study, we have only conducted a very minimal investigation into the different seeds and hyper-parameters of the model we use. Figure \ref{fig:200-200_astronomaly_comparison} shows the standard deviation in our performance when comparing an ensemble of \texttt{AnomalyMatch} models to \texttt{Astronomaly}. As shown there, performance may vary on the anomalies detected dependent on the random seed used. This is not unusual in the context of using neural networks. Further studies are required to fully investigate the effects of hyper parameters and seeds on the performance of the model, however, for now we can state the model performance is comparable to tested classical methods such as \texttt{Astronomaly}.

\section{Conclusion}\label{conclusion_sect}
In this study, we introduced \texttt{AnomalyMatch}, a robust semi-supervised active learning framework tailored for efficiently identifying anomalies in large-scale image datasets. Our pipeline combines the \texttt{FixMatch}-based semi-supervised learning, EfficientNet as a computational backbone with an interactive GUI to allow for active learning during anomaly detection. We have demonstrated excellent performance across diverse datasets (GalaxyMNIST and miniImageNet), highlighting its capability to effectively operate under severe class imbalance and label scarcity -- conditions frequently encountered in astronomy. By significantly reducing the manual review burden, \texttt{AnomalyMatch} offers substantial potential to streamline anomaly discovery processes, effectively harnessing minimal human input and maximising the efficiency of scarce domain-expert resources. The results presented in this work, along with those in the companion paper~\citep{HSTPaper}, underscore the versatility, broad applicability, and potential of \texttt{AnomalyMatch}. Its ability to handle diverse data formats commonly used in astronomy, combined with a user-friendly GUI and integration within ESA Datalabs -- granting direct access to repositories such as the \textit{Euclid} data volumes -- make it a powerful tool for identifying rare astronomical objects in the era of increasingly data-rich sky surveys.

Looking ahead, several promising opportunities exist to further refine and extend the capabilities of \texttt{AnomalyMatch}. Incorporating explainable AI methods -- such as attention maps, SHapley Additive exPlanations~\citep[SHAP;][]{NIPS2017_7062} values, or feature attribution techniques -- could provide crucial interpretability and insight into model decision-making, increasing trust and facilitating model validation by domain experts. Additionally, extending the model to handle multimodal data -- e.g., integrating image information with complementary spectral or time-series data -- would allow richer contextual analysis, potentially enhancing anomaly detection accuracy in complex scientific scenarios. Further improvements could come from developing adaptive training strategies that dynamically calibrate training steps and parameters in response to real-time performance metrics, thus mitigating overfitting risks identified in our current approach. Finally, establishing curated, dedicated benchmark datasets with expert-validated labels, particularly in astronomy, would significantly support methodological advancements, comparison, and standardisation within the anomaly detection research community. Pursuing these enhancements will solidify \texttt{AnomalyMatch} as a versatile and impactful framework, beneficial not only to astronomy but broadly to scientific domains confronted with similar data challenges.

\section*{Data availability statement}
\noindent All data used in benchmarking is readily available from the websites listed below. The \texttt{AnomalyMatch} algorithm is available on GitHub at \url{https://github.com/esa/AnomalyMatch} and on ESA Datalabs at \url{https://datalabs.esa.int/}. All analysis and tests were run using \texttt{AnomalyMatch} version 1.1.0. For the recreation of our results, please use the version released on GitHub. The sources of benchmarking data are:

\begin{enumerate}
    \item The Galaxy Zoo sample is available on Kaggle in the Galaxy Zoo - The Galaxy Challenge. All images and labels are available at \url{https://www.kaggle.com/c/galaxy-zoo-the-galaxy-challenge}. This dataset is described in \citet{2013MNRAS.435.2835W} and \citet{Lochner2021Astronomaly}.
    \item The GalaxyMNIST dataset is available from \url{https://github.com/mwalmsley/galaxy_mnist}.
    \item The miniImageNet dataset is available from \url{https://www.kaggle.com/datasets/arjunashok33/miniimagenet}.
\end{enumerate}

\section*{Conflict of Interest}
The authors declare no conflict of interest.

\section*{Acknowledgments}
This work made use of AI tools for code generation and writing support. The authors acknowledge valuable feedback received from Pedro Mas Buitrago. DOR acknowledges the support of the European Space Agency Research Fellowship Program.

This research made use of many open-source \texttt{Python} packages and scientific computing packages. These include \texttt{Astropy} \citep{astropy:2013, astropy:2018, astropy:2022}, \texttt{Matplotlib} \citep{Hunter:2007}, \texttt{Pandas} \citep{mckinney-proc-scipy-2010}, \texttt{PyTorch} \citep{NEURIPS2019_9015}, \texttt{Numpy} \citep{harris2020array}, \texttt{scikit-image} \citep{van2014scikit} and \texttt{Pillow} \citep{clark2015pillow}.

\bibliographystyle{aasjournal}   
\bibliography{references}  

\begin{appendix}\label{appendix:definitions}
\section{Mathematical formulation of performance metrics}\label{mathematical_definitions}
AUROC and AUPRC are computed based on the counts of true positives (TP), false positives (FP), true negatives (TN), and false negatives (FN). Let \(t \in [0,1]\) be the classification threshold. Then, AUROC and AUPRC can be defined as:
\begin{equation}
\text{AUROC} = \int_{0}^{1} \text{TPR}(t) \, d\text{FPR}(t)\ ,
\end{equation}
\begin{equation}
\text{AUPRC} = \int_{0}^{1} \text{Precision}(t) \, d\text{Recall}(t)\ ;
\end{equation}
Where:
\begin{equation}
\text{True Positive Rate}(t) = \text{TPR}(t) = \frac{\text{TP}(t)}{\text{TP}(t) + \text{FN}(t)}\ ,
\end{equation}
\begin{equation}
\text{False Positive Rate}(t) = \text{FPR}(t) = \frac{\text{FP}(t)}{\text{FP}(t) + \text{TN}(t)}\ ,
\end{equation}
\begin{equation}
\text{Precision}(t) = \frac{\text{TP}(t)}{\text{TP}(t) + \text{FP}(t)} \ ,
\end{equation}
\begin{equation}
\text{Recall}(t) = \text{TPR}(t)\ .
\end{equation}
The Anomaly Detection Efficiency is defined as the fraction of correctly identified anomalies within a specified percentage of top-scoring samples:
\begin{equation}
\text{Efficiency at } p\% = \frac{\text{\# anomalies in top } p\%\text{ scores}}{\text{Total \# anomalies}} \times 100\%\ .
\end{equation}
\end{appendix}
\end{document}